\documentclass[sigconf]{aamas}  

\AtBeginDocument{%
  \providecommand\BibTeX{{%
    \normalfont B\kern-0.5em{\scshape i\kern-0.25em b}\kern-0.8em\TeX}}}

\usepackage{latexsym}
\usepackage{booktabs}       
\usepackage{nicefrac}       

\usepackage{multirow}
\usepackage{makecell}
\usepackage{amsthm}
\usepackage{amssymb}
\usepackage{arydshln}
\usepackage{wrapfig}
\usepackage{algorithm}
\usepackage{algorithmic}
\usepackage{lipsum}
\usepackage{sidecap}
\usepackage{pifont}
\newcommand{\cmark}{\ding{51}}%
\newcommand{\xmark}{\ding{55}}%

\newcommand{\tnt}{\textit{Task \& Talk}}
\newcommand{\ttnc}{\textit{Task, Talk \& Compete}}

\newcommand{\SUB}[1]{\ENSURE \hspace{-0.15in} \textbf{#1}}

\newcommand{\Qbot}{\textsc{Q-bot}}
\newcommand{\Abot}{\textsc{A-bot}}

\definecolor{gg}{RGB}{15,125,15}
\definecolor{rr}{RGB}{190,45,45}

\usepackage{booktabs}    
\usepackage{flushend} 

\setcopyright{acmcopyright}
\setcopyright{ifaamas}  
\copyrightyear{2020} 
\acmYear{2020} 
\acmDOI{} 
\acmPrice{} 
\acmISBN{} 
\acmConference[AAMAS'20]{Proc.\@ of the 19th International Conference on Autonomous Agents and Multiagent Systems (AAMAS 2020)}{May 9--13, 2020}{Auckland, New Zealand}{B.~An, N.~Yorke-Smith, A.~El~Fallah~Seghrouchni, G.~Sukthankar (eds.)}  

\begin{document}

\title{On Emergent Communication in\\Competitive Multi-Agent Teams}  



%
\author{Paul Pu Liang, Jeffrey Chen, Ruslan Salakhutdinov, Louis-Philippe Morency, Satwik Kottur}
\affiliation{%
  \institution{Carnegie Mellon University}
}
\email{pliang@cs.cmu.edu}

\renewcommand{\shortauthors}{P. P. Liang, J. Chen, R. Salakhutdinov, L.-P. Morency, S. Kottur}
\renewcommand{\shorttitle}{On Emergent Communication in Competitive Multi-Agent Teams}

\begin{abstract}
Several recent works have found the emergence of grounded compositional language in the communication protocols developed by mostly cooperative multi-agent systems when learned end-to-end to maximize performance on a downstream task. However, human populations learn to solve complex tasks involving communicative behaviors not only in fully cooperative settings but also in scenarios where competition acts as an additional external pressure for improvement. In this work, we investigate whether competition for performance from an external, similar agent team could act as a social influence that encourages multi-agent populations to develop better communication protocols for improved performance, compositionality, and convergence speed. We start from \tnt, a previously proposed referential game between two cooperative agents as our testbed and extend it into \ttnc, a game involving two competitive teams each consisting of two aforementioned cooperative agents. Using this new setting, we provide an empirical study demonstrating the impact of competitive influence on multi-agent teams. Our results show that an external competitive influence leads to improved accuracy and generalization, as well as faster emergence of communicative languages that are more informative and compositional.
\end{abstract}



\maketitle


\vspace{-1.5mm}
\section{Introduction}

\begin{figure}[tbp]
\centering
\includegraphics[width=\linewidth]{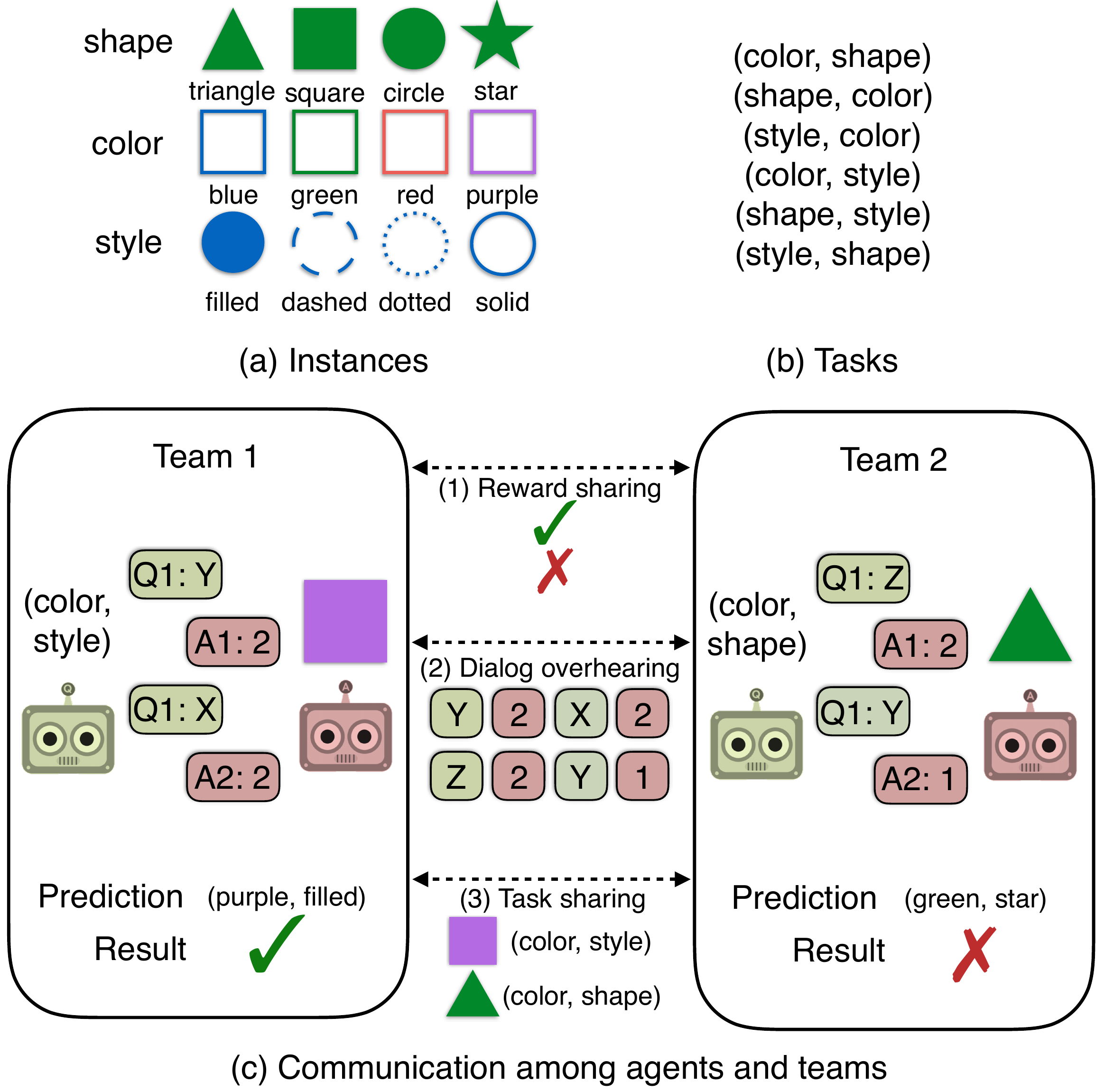}
\caption{We propose the \ttnc \ game involving two competitive teams each consisting of multiple rounds of dialog between \Qbot \ and \Abot. Within each team, \Abot \ is given a target instance unknown to \Qbot \ and \Qbot \ is assigned a task (e.g. find the color and shape of the instance), unknown to \Abot, to uncover certain attributes of the target instance. This informational asymmetry necessitates communication between the two agents via multiple rounds of dialog where \Qbot \ asks questions regarding the task and \Abot \ provides answers using the target instance. We investigate whether \textit{competition for performance} from an external, similar agent team 2 could result in improved compositionality of emergent language and convergence of task performance within team 1. Competition is introduced through three aspects: 1) \textit{reward sharing}, 2) \textit{dialogue overhearing}, and 3) \textit{task sharing}. Our hypothesis is that teams are able to leverage information from the performance of the other team and learn more efficiently beyond the sole reward signal obtained from their own performance. Our findings show that competition for performance with a similar team leads to improved overall accuracy of generalization as well as faster emergence of emergent language.\vspace{-2mm}}
\label{overview}
\end{figure}

The emergence and evolution of languages through human life, societies, and cultures has always been one of the hallmarks of human intelligence~\cite{Nowak8028,reason:Chomsky57a,articleaaa122}. Humans intelligently communicate through language to solve multiple real-world tasks involving vision, navigation, reasoning, and learning~\cite{Nowak2000TheEO,Vogt2005TheEO}. Language emerges naturally for us humans in both individuals through inner speech~\cite{BADDELEY197447,doi:10.1002/9781119132363.ch16,inner} as well as in groups through grounded dialog in both cooperative and competitive settings~\cite{Smith:2003:ILF:963725.963729,Kirby10681,Kirby2015CompressionAC}. As a result, there has been a push to build artificial intelligence models that can effectively communicate their own intentions to both humans as well as other agents through language and dialog grounded in vision~\cite{DBLP:journals/corr/abs-1808-10696,DBLP:conf/aaai/PhamLMMP19,DBLP:journals/corr/abs-1906-02125,multistage} and navigation~\cite{DBLP:journals/corr/abs-1711-11543}.

\begin{figure*}[tbp]
\centering
\includegraphics[width=\linewidth]{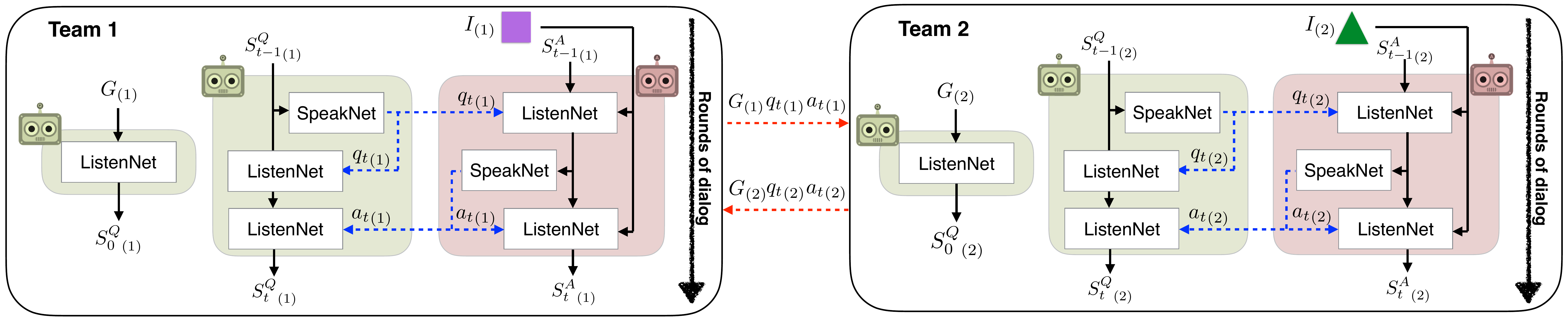}
\caption{Detailed policy networks (implemented as conditional recurrent networks) and communication protocols for \Qbot \ and \Abot \ in both teams. At the start, \Abot \ is given a target instance $I$ unknown to \Qbot \ and \Qbot \ is assigned a task $G$ (unknown to \Abot) to uncover certain attributes of the target instance $I$. At each round $t$, \Qbot \ observes state $s^Q_t$ and utters some token $q_t \in V_Q$. \Abot \ observes the history and this new utterance as state $s^A_t$ and utters $a_t \in V_A$. At the final round, \Qbot \ predicts a pair of attribute values $\hat{w}^G = (\hat{w}^G_1, \hat{w}^G_2)$ to solve the task. Policy parameters $\theta_Q$ and $\theta_A$ are updated using the REINFORCE policy gradient algorithm~\cite{williams1992simple} on a positive reward $+R$ if \Qbot's prediction $\hat{w}^G$ matches ground truth $w^G$ and a negative reward of $-10R$ otherwise. Competition is introduced via 1) \textit{reward sharing} ($R_{(2)}$ passed to team 1), 2) \textit{dialog overhearing} (team 1 overhears and conditions on ${q_t}_{(2)}, {a_t}_{(2)}$), and 3) \textit{task sharing} (team 1 knows about $I_{(2)}$ and $G_{(2)}$). Blue dotted lines represent (cooperative) dialog within a team and red dotted lines represent (competitive) dialog across teams.}
\label{detail}
\end{figure*}

A recent line of work has applied reinforcement learning techniques~\cite{williams1992simple} for end-to-end learning of communication protocols between agents situated in virtual environments~\cite{DBLP:journals/corr/BordesW16,DBLP:journals/corr/LazaridouPB16b} and found the emergence of grounded, interpretable, and compositional symbolic language in the communication protocols developed by the agents~\cite{DBLP:journals/corr/abs-1902-07181,lazaridou2018emergence}. To bring this closer to how natural language evolves within communities, we observe that human populations learn to solve complex tasks involving communicative behaviors not only in fully cooperative settings but also in scenarios where competition acts as an additional external pressure for improvement~\cite{compete,Mobbs9866}. Furthermore, recent studies have also provided support for external competition in behavioral economics~\citep{6706709} and evolutionary biology~\citep{Christiansen1990,doi:10.1111/j.1420-9101.2010.02114.x,fasfas}. Inspired by the emergence of human behavior and language in competitive settings, the goal of our paper is to therefore understand how social influences like competition affects emergent language in multi-agent communication.

In this work, we conceptually distinguish two types of competition: 1) constant sum competition, where agents are competing for a finite amount of resources and thus a gain for a group of agents results in a loss for another~\cite{DBLP:journals/corr/LoweWTHAM17,Bacharach1991}, and 2) variable sum or non-constant sum competition, where competition is inherent but both mutual gains and mutual losses of power are possible~\cite{doi:10.1080/2158379X.2012.659865,doi:10.1177/002224376700400209}. For the sake of simplicity, we call these \emph{competition for resources} and \emph{competition for performance} respectively and focus on the latter. In this setting, each individual (or team) performs similar tasks with access to individual resources but is motivated to do better due to the external pressure from seeing how well other individuals (or teams) are performing. We hypothesize that \emph{competition for performance} from an external, similar agent team could act as a social influence that encourages multi-agent populations to develop better communication protocols for improved performance, compositionality, and convergence.

To investigate this hypothesis, we extend the \tnt\ referential game between two cooperative agents~\cite{DBLP:journals/corr/KotturMLB17} into \ttnc, a game involving two competitive teams each consisting of the aforementioned two cooperative agents (Figure~\ref{overview}). At a high level, \tnt{} requires the two cooperative agents (\Qbot{} and \Abot{}) to solve a \textit{task} by interacting with each other via \textit{dialog}, at the end of which a \textit{reward} is assigned to measure their performance. In such a setting, we introduce competition for performance between two teams of \Qbot{} and \Abot{} through three aspects: 1) \textit{reward sharing}, where we modify the reward structure so that a team can assess their performance relative to the other team, 2) \textit{dialog overhearing}, where we modify the agent's policy networks such that they can overhear the concurrent dialog from the other team, and 3) \textit{task sharing}, where teams gain additional information about the tasks given to the opposing team. Using this new setting, we provide an empirical study demonstrating the impact of competitive influence on multi-agent dialog games. Our results show that external competitive influence leads to an improved accuracy and generalization, as well as faster emergence of communicative languages that are more informative and compositional.

\vspace{-1.5mm}
\section{The \tnt\ Game}

The \tnt\ benchmark is a cooperative reference game proposed to evaluate the emergence of language in multi-agent populations with imperfect information~\cite{DBLP:journals/corr/KotturMLB17,DBLP:journals/corr/abs-1904-09067}. \tnt\ takes place between two agents \Qbot\ (which is tasked to ask questions) and \Abot\ (which is tasked to answer questions) in a synthetic world of instances comprised of three attributes: color, style, and shape. At the start of the game, \Abot\ is given a target instance $I$ unknown to \Qbot\ and \Qbot\ is assigned a task $G$ (unknown to \Abot) to uncover certain attributes of the target instance $I$. This informational asymmetry necessitates communication between the two agents via multiple rounds of dialog where \Qbot\ asks questions regarding the task and \Abot\ provides answers using the target instance. At the end of the game, \Qbot\ uses the information conveyed from the dialog with \Abot\ to solve the task at hand. Both agents are given equal rewards based on the accuracy of \Qbot's prediction.\footnote{detailed review of \tnt\ in the appendix.}

\begin{figure}
    \begin{minipage}{\linewidth}
    \begin{algorithm}[H]
    \caption{Training a single team of cooperative agents.}
    \begin{algorithmic}[1]
        \SUB{\textsc{CooperativeTrain}:}
            \STATE Given \Qbot\ params $\theta_Q$ and \Abot\ params $\theta_A$.
            \FOR{$(I,G)$ in each batch}
                \FOR{communication round $t = 1, ..., T$}
                    \STATE $s^Q_t = [G, q_1, a_1, \dots, q_{t-1}, a_{t-1}]$
                    \STATE $q_t = \pi_Q(q_t | s^Q_t; \theta_Q)$
                    \STATE $s^A_t = [I, q_1, a_1, \dots, q_{t-1},a_{t-1},q_t]$
                    \STATE $a_t = \pi_A(a_t | s^A_t; \theta_A)$
                \ENDFOR
                \STATE $\hat{w}^G = (\hat{w}^G_1, \hat{w}^G_2) = \pi_G (g | s^Q_T; \theta_Q)$
                \STATE $J(\theta_Q,\theta_A) = \mathbb{E}_{\pi_Q,\pi_A} \left[ \mathcal{R} (\hat{w}^G, w^G) \right]$.
                \STATE ${\theta_Q} = {\theta_Q} + \eta \nabla_{{\theta_Q}_{}} J({\theta_Q}_{}, {\theta_A}_{})$.
                \STATE ${\theta_A} = {\theta_A} + \eta \nabla_{{\theta_A}_{}} J({\theta_Q}_{}, {\theta_A}_{})$.\\
                \qquad $\triangleright$ Update $\theta_Q, \theta_A$ using estimated $\nabla J(\theta_Q,\theta_A)$.
            \ENDFOR
    \end{algorithmic}
    \end{algorithm}
    \end{minipage}
\end{figure}

\textbf{Base States and Actions:} Each agent begins by observing its specific input: task $G$ for \Qbot\ and instance $I$ for \Abot. The game proceeds as a dialog over multiple rounds $t = 1, ..., T$. We use lower case characters (e.g. $s_t^Q$) to denote token symbols and upper case $S_t^Q$ to denote corresponding representations. \Qbot\ is modeled with three modules -- speaking, listening, and prediction. At round $t$, \Qbot\ stores an initial state representation $S^Q_{t-1}$ from which it conditionally generates output utterances $q_t\in V_Q$ where $V_Q$ is the vocabulary of \Qbot. $S^Q_{t-1}$ is updated using answers $a_t$ from \Abot\ and is used to make a prediction $\hat{w}_G$ in the final round. \Abot\ is modeled with two modules -- speaking and listening. \Abot\ encodes instance $I$ into its initial state $S_t^A$ from which it conditionally generates output utterances $a_t\in V_A$ where $V_A$ is the vocabulary of \Qbot. $S_t^A$ is updated using questions $q_t$ from \Qbot. 

In more detail, \Qbot\ and \Abot\ are modeled as \textbf{stochastic policies} $\pi_Q(q_t{\mid}s^Q_t; \theta_Q)$ and $\pi_A(a_t{\mid}s^A_t; \theta_A)$ implemented as recurrent networks~\cite{Jain:1999:RNN:553011,hochreiter1997long}. At the beginning of round $t$, \Qbot\ observes state $s^Q_t = [G, q_1, a_1, \dots, q_{t-1}, a_{t-1}]$ representing the task $G$ and the dialog conveyed up to round $t-1$. \Qbot\ conditions on state $s^Q_t$ and utters a question represented by some token $q_t \in V_Q$. \Abot\ also observes the dialog history and this new utterance as state $s^A_t = [I, q_1, a_1, \dots, q_{t-1},a_{t-1},q_t]$. \Abot\ conditions on this state $s^A_t$ and utters an answer $a_t \in V_A$. At the final round, \Qbot\ predicts a pair of attributes $\hat{w}^G = (\hat{w}^G_1, \hat{w}^G_2)$ using a network $\pi_G (g | s^Q_T; \theta_Q)$ to solve the task.

\textbf{Learning the Policy}: \Qbot\ and \Abot\ are trained to cooperate by maximizing a shared objective function that is determined by whether \Qbot\ is able solve the task at hand. \Qbot\ and \Abot\ receive an identical \textbf{base reward} of $R$ if \Qbot's prediction $\hat{w}^G$ matches ground truth $w^G$ and a negative reward of $-10R$ otherwise. $R$ is a hyperparameter that affects the rate of convergence. 
\begin{align}
    J(\theta_Q,\theta_A) = \mathbb{E}_{\pi_Q,\pi_A} \left[ \mathcal{R} (\hat{w}^G, w^G) \right]
\end{align}
where $\mathcal{R}$ is a reward function. To train these agents, we learn policy parameters $\theta_Q$ and $\theta_A$ that maximize the expected reward $J(\theta_Q,\theta_A)$ using the REINFORCE policy gradient algorithm~\cite{williams1992simple}. The expectation of policy gradients for $\theta_Q$ and $\theta_A$ are given by
\begin{align}
    \nabla_{\theta_Q} J(\theta_Q,\theta_A) &= \mathbb{E}_{\pi_Q,\pi_A} \left[ \mathcal{R} (\hat{w}^G, w^G) \nabla_{\theta_Q} \log \pi_Q \left( q_t | s_{t}^Q ; \theta_Q \right) \right],\\
    \nabla_{\theta_A} J(\theta_Q,\theta_A) &= \mathbb{E}_{\pi_Q,\pi_A} \left[ \mathcal{R} (\hat{w}^G, w^G) \nabla_{\theta_A} \log \pi_A \left( a_t | s_{t}^A ; \theta_A \right) \right].
\end{align}
These expectation are approximated by sample averages across object instances and tasks in a batch, as well as across dialog rounds for a given object instance and task. Using the estimated gradients, the parameters $\theta_Q$ and $\theta_A$ are updated using gradient-based methods in an alternating fashion until convergence. We summarize the procedure for cooperative training for a single team of agents in Algorithm 1, which we call \textsc{CooperativeTrain}.

\begin{figure}
    \begin{minipage}{\linewidth}
    \begin{algorithm}[H]
    \caption{Training two teams to compete against each other.}
    \begin{algorithmic}[1]
        \SUB{\textsc{CompetitiveTrain}:}
            \STATE Given: \Qbot$_{(j)}$ params ${\theta_Q}_{(j)}$; \Abot$_{(j)}$ params ${\theta_A}_{(j)}$, for $j = 0,1$.
            \FOR{$(I,G)$ in each batch}
                \FOR{communication round $t = 1, ..., T$}
                    \FOR{$j = 0,1$}
                        \STATE $j' = 1 - j$ \qquad $\triangleright$ Index of the other team
                        \STATE ${s^Q_t}_{(j)} = [G_{(j)}, {q_1}_{(j)}, {a_1}_{(j)}, \dots, {q_{t-1}}_{(j)}, {a_{t-1}}_{(j)}]$
                        \IF{\underline{Dialog Overhearing}}
                            \STATE ${s^Q_t}_{(j)} += [{q_1}_{(j')}, {a_1}_{(j')}, \dots, {q_{t-1}}_{(j')}, {a_{t-1}}_{(j')}]$\\
                            \qquad $\triangleright$ Overhear dialog from the other team
                        \ENDIF
                        \IF{\underline{Task Sharing}}
                        \STATE \ ${s^Q_t}_{(j)} += [G_{(j')}]$
                        \ENDIF
                        \STATE ${q_t}_{(j)} = \pi_Q(q_t | {s^Q_t}_{(j)}; {\theta_Q}_{(j)})$
                        \STATE (and symmetrically for ${s^A_t}_{(j)}$ of \Abot$_{(j)}$)
                    \ENDFOR
                \ENDFOR
                \STATE Compute ${\hat{w}^G} = ({\hat{w}^G_1}, {\hat{w}^G_2}) = \pi_G (g | {s^Q_T}; {\theta_Q})$ independently for each team.
                \IF{\underline{Reward Sharing}}
                    \STATE $J({\theta_Q}_{(j)},{\theta_A}_{(j)}) = \mathbb{E}_{\pi_Q,\pi_A} \left[ \mathcal{R}_{\textrm{shared}} ({\hat{w}^G}_{(j)}, {w^G}_{(j)}) \right]$, $j = 0,1$.\\
                    \qquad $\triangleright$ Compute $J()$ jointly using predictions of \textit{both} teams
                \ELSE 
                    \STATE $J({\theta_Q}_{(j)},{\theta_A}_{(j)}) = \mathbb{E}_{\pi_Q,\pi_A} \left[ \mathcal{R} ({\hat{w}^G}_{(j)}, {w^G}_{(j)}) \right]$, $j = 0,1$.\\
                    \qquad $\triangleright$ Compute $J()$ independently for \textit{each} team
                \ENDIF
                \STATE ${\theta_Q}_{(j)} = {\theta_Q}_{(j)} + \eta \nabla_{{\theta_Q}_{(j)}} J({\theta_Q}_{(j)}, {\theta_A}_{(j)})$, $j = 0,1$.\\
                \STATE ${\theta_A}_{(j)} = {\theta_A}_{(j)} + \eta \nabla_{{\theta_A}_{(j)}} J({\theta_Q}_{(j)}, {\theta_A}_{(j)})$, $j = 0,1$.\\
                \qquad $\triangleright$ Update parameters using estimated $\nabla J()$.
            \ENDFOR
            
        \SUB{\textsc{FullTrain}:}
            \FOR{$epoch = 1, ..., \textrm{MAX}$}
                \STATE \textsc{CooperativeTrain} agents within team 1.
                \STATE \textsc{CooperativeTrain} agents within team 2.
                \STATE \textsc{CompetitiveTrain} agents across teams 1 and 2.
            \ENDFOR
    \end{algorithmic}
    \end{algorithm}
    \end{minipage}
\end{figure}

\vspace{-1.5mm}
\section{The \ttnc\ Game}

We modify the \tnt\ benchmark into the \ttnc\ game (Figure~\ref{detail}). Our setting now consists of two teams of agents: \Qbot$_{(1)}$ and \Abot$_{(1)}$ belonging to team 1 and \Qbot$_{(2)}$ and \Abot$_{(2)}$ belonging to team 2.
Similar to the traditional \tnt\ game, the agents \Qbot$_{(1)}$ and \Abot$_{(1)}$ in team 1 cooperate and communicate to solve a task, and likewise for the agents in team 2.

The \ttnc\ game begins with two target instances $I_{(1)}$ and $I_{(2)}$ presented to \Qbot$_{(1)}$ and \Qbot$_{(2)}$ respectively, and two tasks $G_{(1)}$ and $G_{(2)}$ presented to \Abot$_{(1)}$ and \Abot$_{(2)}$ respectively. Within a team, we largely follow the setting in \tnt\ as described in the previous section. A \textit{team} consists of a pair agents \Qbot\ and \Abot\ cooperating in a partially observable world to solve task $G$ given instance $I$. The key difference is that agents in one team are \textit{competing} against those in the other team. In the following subsections, we explain the various sources of competition that we introduce into the game and the modified training procedure for teams of agents.

We use subscripts to index the rounds and subscripts in parenthesis to index which team the agents belong to (i.e. ${s_t^Q}_{(1)}$). We drop the team subscript if it is clear from the context (i.e. same team). Note that for grounding to happen across teams (i.e. team 1 to understand team 2 during dialog overhearing and vice-versa), the teams \textit{share vocabularies sizes} (i.e. $V_Q$ is shared by \Qbot$_{(1)}$ and \Qbot$_{(2)}$). They do not share vocabularies since both teams train differently but they have access to the same number of symbols.

\vspace{-1.5mm}
\subsection{Sources of Competition}
\label{compete}

When the two teams do not share any information and are trained completely independently, the \ttnc\ game reduces to (two copies of) the \tnt\ game. Therefore, information sharing across the teams is necessary to introduce competition. In the following section we highlight the information sharing that can happen in \ttnc\ and the various sources of competition that can subsequently arise.

\textbf{Reward Sharing (RS):} We modify the reward structure so that a team can assess their performance relative to other teams. Starting with a base reward of $R$ in the fully cooperative setting, we modify this reward into the competitive setting in the following Table:

\vspace{2mm}
\begin{center}
\begin{tabular}{c|c|c}
& Team 2 \textcolor{gg}\cmark & Team 2 \textcolor{rr}\xmark \\ \hline
Team 1 \textcolor{gg}\cmark & $(+R,+R)$ & $(+R,-100R)$ \\
Team 1 \textcolor{rr}\xmark & $(-100R,+R)$ & $(-10R,-10R)$
\end{tabular}
\end{center}
\vspace{2mm}

When both teams get the same result this reduces to the base reward setting: $+R$ for correct answers and $-10R$ for wrong ones. When there is asymmetry in performance across the two teams, the correct team gains a reward of $+R$ while the incorrect team suffers a larger penalty of $-100R$. This encourages teams to compete and assess their performance relative to the other team.

\textbf{Dialog Overhearing (DO):} Overhearing the conversations of another team is a common way to get secret information about the tactics, knowledge, and progress of one's competitors. In a similar fashion, we modify the agent's policy networks such that they can now overhear the concurrent dialog from other teams. Take \Qbot$_{(1)}$ and \Abot$_{(1)}$ in team 1 for example. At round $t$, \Qbot$_{(1)}$ now observes state
\begin{align}
\label{qbot}
\nonumber &{s^Q_t}_{(1)} = [G_{(1)}, {q_1}_{(1)}, {a_1}_{(1)}, {q_1}_{(2)}, {a_1}_{(2)}, \dots, \\
&{q_{t-1}}_{(1)}, {a_{t-1}}_{(1)}, {q_{t-1}}_{(2)}, {a_{t-1}}_{(2)}],
\end{align}
and similarly for \Qbot$_{(2)}$. \Abot$_{(1)}$ observes 
\begin{align}
\label{abot}
\nonumber &{s^A_t}_{(1)} = [I_{(1)}, {q_1}_{(1)}, {a_1}_{(1)}, {q_1}_{(2)}, {a_1}_{(2)}, \dots, \\
&{q_{t-1}}_{(1)}, {a_{t-1}}_{(1)}, {q_{t-1}}_{(2)}, {a_{t-1}}_{(2)}, {q_t}_{(1)}, {q_t}_{(2)}].
\end{align}
and similarly for \Abot$_{(2)}$. We view this as augmenting reward sharing by informing the agents as to why they were penalized: a team can listen to what the other team is communicating and use that to its own advantage. In practice, we define an overhear fraction $\rho$ which determines how often overhearing occurs during the training epochs. $\rho$ is a hyperparameter in our experiments.

\textbf{Task Sharing (TS):} Finally, we fully augment both reward sharing and dialog overhearing with task sharing, where the two sets of instances and tasks are known to both teams (i.e. $G_{(2)}$ is added to Equation~\ref{qbot} and $I_{(2)}$ is added to Equation~\ref{abot}). Task sharing adds an additional level of grounding: one team can now ground the overheard dialog in a specific task and compare their performance to that of the other team. Overall, Algorithm 2 summarizes the procedure for training teams of agents in a competitive setting using the aforementioned three sources of competition. We call the resulting algorithm \textsc{CompetitiveTrain}.

\textbf{Asynchronous Training:} The baseline from~\citet{DBLP:journals/corr/KotturMLB17} only evaluates the performance of a single team at test time. For fair comparison with this baseline, we evaluate performance with a single team 1 as well. In the case of dialog overhearing and task sharing, we replace the overheard symbols and tasks from task 2 with zeros during evaluation. This removes the confounding explanation that improved performance is due to having more information from team 2 during testing.

To prevent data mismatch during training and testing~\cite{Quionero-Candela:2009:DSM:1462129,Glorot:2011:DAL:3104482.3104547}, we use a three-stage training process. During stage (1), \Qbot$_{(1)}$ and \Abot$_{(1)}$ are trained to cooperate, independent of team 2. All information shared from team 2 is passed to team 1 as zeros. During stage (2), \Qbot$_{(2)}$ and \Abot$_{(2)}$ are trained to cooperate, and in stage (3), both teams are trained together with reward sharing and/or dialog overhearing and/or task sharing. These three stages are repeated until convergence. Intuitively, this procedure means that the agents within each team must learn how to cooperate with each other in addition to competing with the other team. The final algorithm which takes into account asynchronous training both within and across teams is shown in Algorithm 2, which we call \textsc{FullTrain}. FullTrain is the final algorithm used for training both teams of agents.

Due to how we model competition in our setting (either through reward sharing or task and dialog overhearing), our hypothesis is that teams are able to leverage information from the performance of the other team and learn more efficiently beyond the signal obtained solely from their performance.

\begin{table*}[t]
\fontsize{9}{11}\selectfont
\setlength\tabcolsep{3.0pt}
\centering
\begin{tabular}{l | l | c c c | c c | c c}
\Xhline{3\arrayrulewidth}
\multirow{2}{*}{Type} & \multirow{2}{*}{Method} & \multirow{2}{*}{RS} & \multirow{2}{*}{DO} & \multirow{2}{*}{TS} & \multicolumn{2}{c|}{Train Acc (\%)} & \multicolumn{2}{c}{Test Acc (\%)}\\
& & & & & Winning Team & Losing Team & Winning Team & Losing Team\\
\Xhline{0.5\arrayrulewidth}
\multirow{4}{*}{\makecell{{\color{blue}{Cooperative}}\\ {\color{blue}{baselines}}}} & Coop, base~\cite{DBLP:journals/corr/KotturMLB17}   & \textcolor{rr}\xmark & \textcolor{rr}\xmark & \textcolor{rr}\xmark & $88.5 \pm 11.6$ & - & $45.6 \pm 18.9$ & -\\
& Coop, rewards~\cite{Grzes:2008:MRL:1429293.1429335,Grzes:2017:RSE:3091125.3091208}  & \textcolor{rr}\xmark & \textcolor{rr}\xmark & \textcolor{rr}\xmark & $87.0 \pm 13.7$ & - & $49.7 \pm 22.9$ & - \\
& Coop, params   & \textcolor{rr}\xmark & \textcolor{rr}\xmark & \textcolor{rr}\xmark & $85.5 \pm 14.6$ & - & $53.3 \pm 26.2$ & - \\
& Coop, double    & \textcolor{rr}\xmark & \textcolor{rr}\xmark & \textcolor{rr}\xmark & $91.4 \pm 12.0$ & $74.6 \pm 16.6$ & $57.8 \pm 28.5$ & $38.3 \pm 27.0$\\
\Xhline{0.5\arrayrulewidth}
\multirow{7}{*}{\makecell{{\color{rr}{Competitive}}\\ {\color{rr}{methods}}}} & Comp, TS & \textcolor{rr}\xmark & \textcolor{rr}\xmark & \textcolor{gg}\cmark & $94.4 \pm 3.0$ & $80.4 \pm 13.3$ & $53.1 \pm 19.5$ & $41.7 \pm 25.7$\\
& Comp, DO     & \textcolor{rr}\xmark & \textcolor{gg}\cmark & \textcolor{rr}\xmark & $96.8 \pm 5.4$ & $71.4 \pm 14.2$ & $65.7 \pm 26.1$ & $23.1 \pm 10.7$\\
& Comp, DO+TS     & \textcolor{rr}\xmark & \textcolor{gg}\cmark & \textcolor{gg}\cmark & $98.6 \pm 2.0$ & $75.8 \pm 12.1$ & $\mathbf{75.8 \pm 17.9}$ & $41.1 \pm 26.0$\\
& Comp, RS  & \textcolor{gg}\cmark & \textcolor{rr}\xmark & \textcolor{rr}\xmark & $99.5 \pm 1.3$ & $79.6 \pm 12.2$ & $68.5 \pm 19.4$ & $43.8 \pm 20.9$\\
& Comp, RS+TS     & \textcolor{gg}\cmark & \textcolor{rr}\xmark & \textcolor{gg}\cmark & $99.5 \pm 1.4$ & $66.6 \pm 10.3$ & $68.9 \pm 16.9$ & $26.7 \pm 11.3$\\
& Comp, RS+DO    & \textcolor{gg}\cmark & \textcolor{gg}\cmark & \textcolor{rr}\xmark & $\mathbf{100.0 \pm 0.0}$ & $63.9 \pm 12.8$ & $\mathbf{78.3 \pm 10.7}$ & $28.3 \pm 14.4$\\
& Comp, RS+DO+TS      & \textcolor{gg}\cmark & \textcolor{gg}\cmark & \textcolor{gg}\cmark & $98.8 \pm 2.4$ & $78.9 \pm 10.9$ & $\mathbf{77.2 \pm 16.5}$ & $28.9 \pm 21.8$\\
\Xhline{3\arrayrulewidth}
\end{tabular}
\vspace{2mm}
\caption{Train and test accuracies measured across teams trained in various cooperative and competitive settings. All cooperative baselines are in shades of {\color{blue} blue} and competitive teams are in {\textcolor{rr}{red}}. RS: reward sharing, DO: dialog overhearing, TS: task sharing. Accuracies are reported separately for winning and losing teams with best accuracies for winning teams in \textbf{bold}. Winning teams in competitive settings display faster convergence and improved performance.\vspace{-4mm}}
\label{res}
\end{table*}

\vspace{-1.5mm}
\section{Experimental Setup}

We present our experimental results and observations on the effect of competitive influence on both final task performance and emergence of grounded, interpretable, and compositional language. Our experimental testbed is the \ttnc\ game. We implement a variety of different algorithms spanning cooperative baselines and competitive methods which we will detail in the following subsections.\footnote{Our data and models are publicly released at \url{https://github.com/pliang279/Competitive-Emergent-Communication}.}

\vspace{-1.5mm}
\subsection{Cooperative Baselines}

These are baseline methods that involve only cooperative team (or teams) of agents. These baselines test for various confounders of our experiments (e.g. improved performance due to increase in the number of parameters).

(1) \textsc{Coop, base:} a single team, fully cooperative setting with reward structure $(+R,-10R)$, which is the baseline from~\cite{DBLP:journals/corr/KotturMLB17}.

(2) \textsc{Coop, rewards:} a single team, fully cooperative setting with reward structure $(+R,-100R)$ adjusted for our ``extreme'' reward setting that is more strict in penalizing incorrect answers. This baseline ensures that our improvements are not simply due to better reward shaping~\cite{Grzes:2008:MRL:1429293.1429335,Grzes:2017:RSE:3091125.3091208}.

(3) \textsc{Coop, params:} a single team, fully cooperative setting with (roughly) double the number of parameters as compared to the baseline. This ensures that the improvement in performance is not due to an increase in the number of parameters during dialog overhearing or task sharing. This baseline is obtained by increasing the LSTM hidden size for question and answer bots such that the total number of parameters match the competitive settings with roughly double the number of parameters.

(4) \textsc{Coop, double:} two teams each trained independently without sharing any information. At test time, the team with the higher validation score is used to evaluate performance. This baseline ensures that our improvements are not simply due to double the chance of beginning with a better random seed or luckier training.

\vspace{-1.5mm}
\subsection{Competitive Methods}

The following methods introduce an extra level of competition between teams on top of cooperation within teams. Competitive behavior is encouraged via combinations of reward sharing, dialog overhearing, and task sharing.

(1) \textsc{Comp, TS:} two competitive teams with task sharing.

(2) \textsc{Comp, DO:} two competitive teams with dialog overhearing.

(3) \textsc{Comp, DO+TS:} two competitive teams with dialog overhearing and task sharing.

(4) \textsc{Comp, RS:} two competitive teams with reward sharing.

(5) \textsc{Comp, RS+TS:} two competitive teams with reward sharing and task sharing.

(6) \textsc{Comp, RS+DO:} two competitive teams with reward sharing and dialog overhearing.

(7) \textsc{Comp, RS+DO+TS:} two competitive teams with reward sharing, dialog overhearing, and task sharing.

\textbf{Hyperparameters:} We perform all experiments with the same hyperparameters (except LSTM hidden dimension which is fixed at 100 but increased to 150 for the \textsc{Coop, params} setting to experiment with an increase in the number of parameters). We set the reward multiplier $R = 100$, overhear fraction $\rho=0.5$, and vocabulary sizes of \Qbot\ and \Abot\ to be $|V_Q|=3$ and $|V_A|=4$ respectively. Following~\citet{DBLP:journals/corr/KotturMLB17}, we set \Abot$_{(1)}$ and \Abot$_{(2)}$ to be memoryless to ensure consistent \Abot\ grounding across rounds which is important for generalization and compositional language. All other parameters follow those in~\citet{DBLP:journals/corr/KotturMLB17}.

\textbf{Metrics:} In addition to evaluating the train and test accuracies~\cite{DBLP:journals/corr/KotturMLB17}, we would also like to investigate the impact of competitive pressure on the emergence of language among agents. We measure \textbf{Instantaneous Coordination (IC)}~\cite{jaques2019intrinsic} defined as the mutual information between one agent's message and \textit{another} agent's action, i.e. $\textrm{MI} ({a_t}_{(j)}, {\hat{w}}^G_{i(j)})$. Higher IC implies that \Qbot's action depends more strongly on \Abot's message (and vice versa), which is in turn indicative that messages are used in positive manner to signal actions within a team. Although~\citet{DBLP:journals/corr/abs-1903-05168} mentioned other metrics such as speaker consistency~\cite{jaques2019intrinsic,DBLP:journals/corr/abs-1809-00549} and entropy, we believe that IC is the most suited for question answering dialog tasks where responding to \textit{another} agent's messages is key. We also measured several other metrics that were recently proposed to measure how informative a language is with respect to the agent's actions~\cite{DBLP:journals/corr/abs-1903-05168}: \textbf{Speaker Consistency (SC)} measures the mutual information between an agent's message and its own future action: $\textrm{MI} ({q_t}_{(j)}, {\hat{w}}^G_{i(j)})$ and \textbf{Entropy (H)} which measures the entropy of an agent's sequence of outgoing messages.

\textbf{Training Details:} For cooperative baselines, we set the maximum number of epochs to be 100,000 and stop training early when training accuracy reaches $100\%$. For competitive baselines, we also set the maximum number of epochs to be 100,000 and stop training early when \textit{the first team} reaches a training accuracy of $100\%$. This team is labeled as the \textit{winning} team and is the focus of our experiments. The other \textit{losing} team can be viewed as an auxiliary team that helps the performance of the winning team. All experiments are repeated $10$ times with randomly chosen random seeds. Results are reported as average $\pm$ standard deviation over the $10$ runs. Implementation details are provided in the appendix.

\vspace{-1.5mm}
\section{Results and Analysis}

We study the effect of competition between teams on 1) the generalization abilities of the agents in new environments during test-time, 2) the rate of convergence of train and test accuracies during training, and 3) the emergence of informative communication protocols between agents when solving the \ttnc\ game.

\vspace{-1.5mm}
\subsection{Qualitative Results}

We begin by studying the effect of competition on the performance of agents in the \ttnc\ game. The teams are trained in various cooperative and competitive settings. We aggregate scores for both winning and losing teams as described in the training details above. These results are summarized in Table~\ref{res}. From our results, we draw the following general observations regarding the generalization capabilities of the trained agents:

(1) \textbf{Sharing messages via overhearing dialog improves generalization performance:} Dialog overhearing contributes the most towards improvement in test accuracy, from below $60\%$ in the baselines without to $75.8\%$ with dialog overhearing. We believe this is because dialog overhearing transmits the most amount of information to the other team, as compared to a single scalar in reward sharing or a single image in task sharing.

(2) \textbf{Composing sources of competition improves performance:} While dialog overhearing on its own displays strong improvements in test performance, we found that composing multiple sources of competition improves performance even more. In the \textsc{Comp, RS+DO} setting, the winning team's train accuracy quickly increases to $100\%$ very consistently across all $10$ runs, while test accuracy is also the highest at $78.3\%$. Other settings that worked well included \textsc{Comp, DO+TS} with a winning test accuracy of $75.8\%$, and \textsc{Comp, RS+DO+TS} with a winning test accuracy of $77.2\%$.

(3) \textbf{Increasing competitive pressure increases the gap between winning and losing teams:} Another finding is that as more competitive pressure is introduced, the gap between winning and losing teams increases. The winning team increasingly performs better, especially in train accuracy, and the losing team increasingly performs worse than the single team cooperative baseline (i.e. lower than $\sim50\%$). This confirms our hypothesis that the losing team acts as an \textit{auxiliary} team that boosts the performance of the winning team at its own expense. Furthermore, winning teams that survive through multiple sources of competition learn better communication protocols that allow them to generalize better to new test environments.

(4) \textbf{Reward shaping does not improve performance:} In both cooperative and competitive settings, one cannot rely solely on reward shaping to improve generalization. The test accuracies are largely similar across various reward settings.

(5) \textbf{Improvement in performance is not due to other factors:} To ensure that the empirical results we observe are not due to confounding factors, we compare the performance in teams of competitive agents with purely cooperative baselines with reward shaping, doubling the number of parameters, and doubling the number of teams. None of these baselines generalize well which shows that the improvement in performance is not due to better rewards, more parameters, or luckier training.

\begin{figure*}[tbp]
\centering
\includegraphics[width=\linewidth]{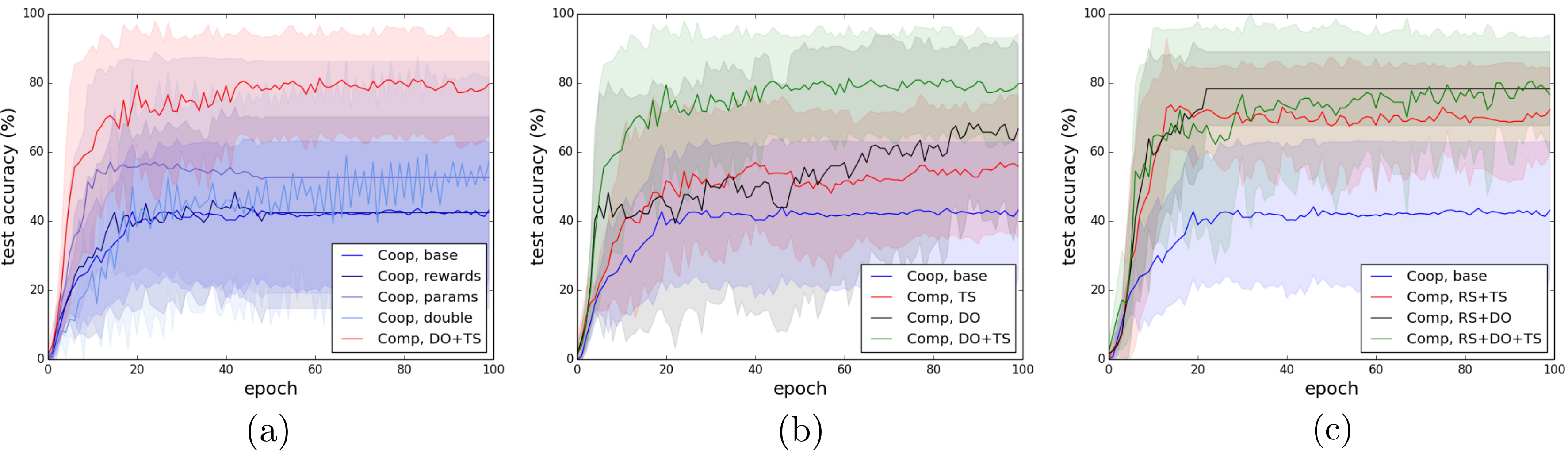}
\caption{We plot the convergence of test accuracy across training epochs for the winning team trained in various cooperative and competitive settings. All cooperative baselines are in shades of {\color{blue} blue} and competitive teams are in {\textcolor{rr}{red}}, {\color{black} black}, and {\textcolor{gg}{green}}. Lines represent the mean across 10 runs and shaded boundaries represent the standard deviations. In (a), we compare the cooperative baselines \textsc{Coop, base}, \textsc{Coop, rewards}, \textsc{Coop, params}, and \textsc{Coop, double} with a well-performing competitive method, \textsc{Comp, DO+TS}. In (b), we compare \textsc{Coop, base} with competitive teams involving Dialog Overhearing (DO) and Task Sharing (TS) (i.e. \textsc{Comp, TS}, \textsc{Comp, DO}, \textsc{Comp, DO+TS}). In (c), we compare \textsc{Coop, base} with competitive teams that additionally incorporate Reward Sharing (RS) (i.e. \textsc{Comp, RS+TS}, \textsc{Comp, RS+DO}, \textsc{Comp, RS+DO+TS}). Winning teams in competitive settings display faster convergence and improved generalization performance in test environments.\vspace{-2mm}}
\label{plots}
\end{figure*}

\vspace{-1.5mm}
\subsection{Rates of Convergence}

We also compare the convergence in test accuracies of teams trained in both fully cooperative and competitive settings. For test accuracies that ended due to early stopping when train accuracy reached $100\%$, we propagate the test accuracy corresponding to the best train accuracy over the remaining epochs. This ensures that the test accuracies over multiple runs are averaged accurately across epochs. We outline these results in Figure~\ref{plots}. We find that in all settings, training teams of competitive agents leads to faster convergence and improved performance as compared to the cooperative baselines. Furthermore, composing multiple sources of competition during training steadily improves performance. Figure~\ref{plots}(b) shows a clear trend that: \textsc{Comp, DO+TS} $>$ \textsc{Comp, DO} $\approx$ \textsc{Comp, TS} $>$ \textsc{Coop, base}. By further adding reward sharing, Figure~\ref{plots}(c) shows that \textsc{Comp, RS+DO+TS} $\approx$ \textsc{Comp, RS+DO} $>$ \textsc{Comp, RS+TS} $>$ \textsc{Coop, base}.

We find that the fastest convergence in training happens in \textit{earlier stages} of training. Winning teams tend to quickly pull ahead of their losing counterparts during earlier stages and losing teams are unable to recover in later stages of training. This observation is also shared in Table~\ref{res} where the gap between winning and losing teams grows as sources of competition are composed. Interestingly, these observations are also mirrored in studies in psychology which argue that competition during childhood is beneficial for early cognitive and social development~\cite{Pepitone1985,doi:10.1177/0895904811428973}.

Finally, another interesting observation is that high-performing winning teams trained in competitive settings tend to display \textit{lower variance} in test time evaluation as compared to their cooperative counterparts, thereby learning to more stable training.

\begin{table}[t]
\fontsize{9}{11}\selectfont
\setlength\tabcolsep{2.0pt}
\centering
\begin{tabular}{l | l | c c}
\Xhline{3\arrayrulewidth}
\multirow{2}{*}{Type} & \multirow{2}{*}{Method} & \multicolumn{2}{c}{IC} \\
& & Winning Team & Losing Team\\
\Xhline{0.5\arrayrulewidth}
\multirow{4}{*}{\makecell{{\color{blue}{Cooperative}}\\ {\color{blue}{baselines}}}} & Coop, base~\cite{DBLP:journals/corr/KotturMLB17}  & $0.675\pm0.099$ & -\\
& Coop, rewards~\cite{Grzes:2008:MRL:1429293.1429335,Grzes:2017:RSE:3091125.3091208}  & $0.646\pm0.050$ & -\\
& Coop, params    & $0.689\pm0.101$ & -\\
& Coop, double    & $0.719\pm0.145$ & $0.691\pm0.153$\\
\Xhline{0.5\arrayrulewidth}
\multirow{7}{*}{\makecell{{\color{rr}{Competitive}}\\ {\color{rr}{methods}}}} & Comp, TS & $0.650\pm0.139$ & $0.592\pm0.128$\\
& Comp, DO       & $0.778\pm0.161$ & $0.757\pm0.179$\\
& Comp, DO+TS    & $\mathbf{0.806\pm0.202}$ & $0.800\pm0.204$\\
& Comp, RS       & $0.793\pm0.165$ & $0.776\pm0.161$\\
& Comp, RS+TS    & $0.726\pm0.207$ & $0.718\pm0.118$\\
& Comp, RS+DO    & $\mathbf{0.814\pm0.154}$ & $0.743\pm0.116$\\
& Comp, RS+DO+TS & $\mathbf{0.834\pm0.203}$ & $0.740\pm0.142$\\
\Xhline{3\arrayrulewidth}
\end{tabular}
\vspace{2mm}
\caption{The Instantaneous Coordination (IC) metric measured across teams trained in various cooperative and competitive settings. All cooperative baselines are in shades of {\color{blue} blue} and competitive teams are in {\textcolor{rr}{red}}. RS: reward sharing, DO: dialog overhearing, TS: task sharing. IC scores are reported separately for winning and losing teams with best accuracies for winning teams in \textbf{bold}. Winning teams in competitive settings perform more informative communication as measured by a higher IC score.\vspace{-6mm}}
\label{ic}
\end{table}

\begin{table}[t]
\fontsize{9}{11}\selectfont
\setlength\tabcolsep{2.0pt}
\centering
\begin{tabular}{l | l | c c c}
\Xhline{3\arrayrulewidth}
Type & Method & $|V_Q|$ & $|V_A|$ & Winning Team \\
\Xhline{0.5\arrayrulewidth}
\multirow{3}{*}{\makecell{{\color{blue}{Cooperative}}\\ {\color{blue}{baselines}}}} & \multirow{3}{*}{Coop, base~\cite{DBLP:journals/corr/KotturMLB17}}  & $3$ & $4$ & $45.6\pm18.9$\\
& & $16$ & $16$ & $26.4\pm5.1$\\
& & $64$ & $64$ & $22.6\pm4.6$\\
\Xhline{0.5\arrayrulewidth}
\multirow{3}{*}{\makecell{{\color{rr}{Competitive}}\\ {\color{rr}{methods}}}} & \multirow{3}{*}{Comp, RS+DO+TS} & $3$ & $4$ & $\mathbf{77.2\pm16.5}$\\
& & $16$ & $16$ & $50.8\pm26.1$\\
& & $64$ & $64$ & $47.5\pm25.2$\\
\Xhline{3\arrayrulewidth}
\end{tabular}
\vspace{2mm}
\caption{Effect of vocabulary size on both cooperative and competitive training. Similar to~\citet{DBLP:journals/corr/KotturMLB17}, we found that test performance is hurt at large vocab sizes, even under competitive training. For the same fixed vocabulary size, we also see consistent improvements using competitive training as compared to the cooperative baselines, suggesting the utility of our approach across different hyperparameter settings s such as vocabulary sizes.\vspace{-6mm}}
\label{vocab}
\end{table}

\begin{table*}[t]
\fontsize{9}{11}\selectfont
\setlength\tabcolsep{3.0pt}
\centering
\begin{tabular}{l | l | c c | c c}
\Xhline{3\arrayrulewidth}
\multirow{2}{*}{Type} & \multirow{2}{*}{Method} & \multicolumn{2}{c|}{SC $(\uparrow)$} & \multicolumn{2}{c}{H $(\downarrow)$}\\
& & Winning Team & Losing Team & Winning Team & Losing Team\\
\Xhline{0.5\arrayrulewidth}
\multirow{4}{*}{\makecell{{\color{blue}{Cooperative}}\\ {\color{blue}{baselines}}}} & Coop, base~\cite{DBLP:journals/corr/KotturMLB17}  & $0.631\pm0.114$ & - & $1.186\pm0.124$ & - \\
& Coop, rewards~\cite{Grzes:2008:MRL:1429293.1429335,Grzes:2017:RSE:3091125.3091208}  & $0.640\pm0.117$ & - & $1.060\pm0.023$ & - \\
& Coop, params    & $0.676\pm0.132$ & - & $1.138\pm0.143$ & -\\
& Coop, double    & $0.683\pm0.150$ & $0.675\pm0.133$ & $1.196\pm0.128$ & $1.210\pm0.131$ \\
\Xhline{0.5\arrayrulewidth}
\multirow{7}{*}{\makecell{{\color{rr}{Competitive}}\\ {\color{rr}{methods}}}} & Comp, TS & $0.610\pm0.123$ & $0.618\pm0.143$ & $1.089\pm0.119$ & $1.107\pm0.154$\\
& Comp, DO       & $0.635\pm0.149$ & $0.647\pm0.185$ & $1.212\pm0.123$ & $1.210\pm0.135$\\
& Comp, DO+TS    & $0.635\pm0.146$ & $0.646\pm0.188$ & $1.215\pm0.133$ & $1.211\pm0.124$\\
& Comp, RS       & $0.677\pm0.149$ & $0.658\pm0.118$ & $1.205\pm0.137$ & $1.207\pm0.137$\\
& Comp, RS+TS    & $0.622\pm0.157$ & $0.596\pm0.143$ & $1.197\pm0.143$ & $1.205\pm0.130$\\
& Comp, RS+DO    & $\mathbf{0.701\pm0.114}$ & $0.670\pm0.073$ & $1.174\pm0.151$ & $1.217\pm0.123$\\
& Comp, RS+DO+TS & $\mathbf{0.679\pm0.219}$ & $0.615\pm0.157$ & $1.197\pm0.164$ & $1.207\pm0.150$\\
\Xhline{3\arrayrulewidth}
\end{tabular}
\vspace{2mm}
\caption{Speaker Consistency (SC) and Entropy (H) metrics measured across teams in all settings. All cooperative baselines are in shades of {\color{blue} blue} and competitive teams are in {\textcolor{rr}{red}}. RS: reward sharing, DO: dialog overhearing, TS: task sharing. SC and H scores are reported separately for winning and losing teams with best accuracies for winning teams in bold.\vspace{-4mm}}
\label{res_supp}
\end{table*}

\vspace{-1.5mm}
\subsection{Emergence of Language}

In addition to generalization performance, we also compare the quality of the communication protocols that emerge between the teams of trained agents. Specifically, we measure the signaling that occurs between agents using the Instantaneous Coordination (IC) metric~\cite{jaques2019intrinsic}. We report these results in Table~\ref{ic} and focus on the IC between \Qbot\ and \Abot\ from the winning team. We observe that IC is highest for the fully competitive setting \textsc{Comp, RS+DO+TS}. Furthermore, by comparing Table~\ref{res} with Table~\ref{ic}, we observe a strong correlation between winning teams that signal clearly with high IC scores and winning teams that perform best on test environments. \textsc{Comp, DO+TS}, \textsc{Comp, RS+DO}, and \textsc{Comp, RS+DO+TS} are the training settings that lead to such winning teams. These observations supports our hypothesis that having external pressure from similar agents encourages the team's \Qbot\ and \Abot\ to coordinate better through emergent language, thereby leading to superior task performance which is another benefit of our proposed competitive training method.

Finally, we find that the learned communication protocol is compositional in the same measure as~\citet{DBLP:journals/corr/KotturMLB17}. For example, \Qbot\ assigns $Y$ to represent tasks (shape, style), (style, shape), and $X$ for (style, color). The small vocabulary size and memoryless \Abot\ means that the messages must compose across entities to generalize at test time to unseen instances. We further note that from the convergence graphs as shown in Figure~\ref{plots}, compositionality in emergent language is achieved faster in competitive settings as compared to the fully cooperative counterparts.

We also experimented with large vocab sizes of $|V_Q| = |V_A| = 16$ and $64$. We reported these results in Table~\ref{vocab}. Similar to~\citet{DBLP:journals/corr/KotturMLB17}, we found that test performance is hurt at large vocab sizes, even under competitive training. Therefore, we set the vocabulary sizes $|V_Q| = 3$ and $|V_A| = 4$ respectively following~\citet{DBLP:journals/corr/KotturMLB17}. With these limited vocabulary sizes, we observed good generalization of the language to new object instances. When using large vocabulary sizes, the agents tend to use every vocabulary symbol to memorize pairs of concepts, e.g. symbol $a$ represents a green circle and symbol $b$ represents a green square, etc. instead of representing compositional concepts e.g. symbol $a$ represents the color green and symbol $b$ represents the shape square etc. The compositional vocabulary learned in the latter case is required for generalization to new pairs of concepts at test-time.

From Table~\ref{vocab}, it is interesting to note that for the same fixed vocabulary size, we also see consistent improvements using competitive training as compared to the cooperative baselines. This further suggests the utility of our approach across different hyperparameter settings. Moreover, it suggests that competitive training approaches are more robust to different hyperparameter settings such as vocabulary sizes.

\vspace{-1.5mm}
\subsection{Speaker Consistency and Entropy}

Here we report the results on two more metrics proposed to measure how informative a language is with respect to the agent's actions~\cite{DBLP:journals/corr/abs-1903-05168}: \textbf{Speaker Consistency (SC)} measures the mutual information between an agent's message and its future action: $\textrm{MI} ({q_t}_{(j)}, {\hat{w}}^G_{i(j)})$ and \textbf{Entropy (H)} which measures the entropy of an agent's sequence of outgoing messages. We show these experimental results in Table~\ref{res_supp}.

In general, competitive teams display a higher speaker consistency score, again showing strong correlation with the best performing teams \textsc{Comp, RS+DO} and \textsc{Comp, RS+DO+TS}. This again implies that the better performing teams trained via competition demonstrate more signaling using their vocabulary. As for entropy, it is hard to interpret this metric~\cite{DBLP:journals/corr/abs-1903-05168}. It is traditionally thought that lower entropy in languages represents more compositionality and efficiency in the way meaning is encoded in language. On one hand, it is also possible for an agent to always to send the same symbol which implies the lowest possible entropy, but these messages are unlikely to be informative. The results show that the entropies across all settings are roughly similar, which we believe imply that the agents are learning communication protocols that are equally complex and rich in nature. However, the improved speaker consistency and instantaneous coordination scores imply that the communication protocols learnt via competition are more informational to the other agents.

\vspace{-1.5mm}
\section{Conclusion}

In this paper, we revisited emergent language in multi-agent teams from the lens of \textit{competition for performance}: scenarios where competition acts as an additional external pressure for improvement. We start from \tnt, a previously proposed referential game between two cooperative agents as our testbed and extend it into \ttnc, a game involving two competitive teams each consisting of cooperative agents. Using our newly proposed \ttnc\ benchmark, we showed that competition from an external team acts as social influence that encourages multi-agent populations to develop more informative communication protocols for improved generalization and faster convergence. Our controlled experiments also show that these results are not due to confounding factors such as more parameters, more agents, and reward shaping. This line of work constitutes a step towards studying the emergence of language from agents that are both cooperative and competitive at different levels. Future work can explore the effect of competitive multi-agent training in various real-world settings as well as the emergence of natural language and multimodal dialog.

\vspace{-1.5mm}
\section{Acknowledgements}

This material is based upon work partially supported by the National Science Foundation (Awards \#1750439, \#1722822), National Institutes of Health, DARPA SAGAMORE HR00111990016, AFRL CogDeCON FA875018C0014, and NSF IIS1763562. Any opinions, findings, and conclusions or recommendations expressed in this material are those of the author(s) and do not necessarily reflect the views of the National Science Foundation, National Institutes of Health, DARPA, and AFRL, and no official endorsement should be inferred. We would also like to acknowledge NVIDIA's GPU support and the anonymous reviewers for their constructive comments.


\bibliographystyle{ACM-Reference-Format}  
\bibliography{sample-aamas20}  


\begin{thebibliography}{00}


\ifx \showCODEN    \undefined \def \showCODEN     #1{\unskip}     \fi
\ifx \showDOI      \undefined \def \showDOI       #1{#1}\fi
\ifx \showISBNx    \undefined \def \showISBNx     #1{\unskip}     \fi
\ifx \showISBNxiii \undefined \def \showISBNxiii  #1{\unskip}     \fi
\ifx \showISSN     \undefined \def \showISSN      #1{\unskip}     \fi
\ifx \showLCCN     \undefined \def \showLCCN      #1{\unskip}     \fi
\ifx \shownote     \undefined \def \shownote      #1{#1}          \fi
\ifx \showarticletitle \undefined \def \showarticletitle #1{#1}   \fi
\ifx \showURL      \undefined \def \showURL       {\relax}        \fi
\providecommand\bibfield[2]{#2}
\providecommand\bibinfo[2]{#2}
\providecommand\natexlab[1]{#1}
\providecommand\showeprint[2][]{arXiv:#2}

\bibitem[\protect\citeauthoryear{Alderson-Day and Fernyhough}{Alderson-Day and
  Fernyhough}{2015}]%
        {inner}
\bibfield{author}{\bibinfo{person}{Ben Alderson-Day} {and}
  \bibinfo{person}{Charles Fernyhough}.} \bibinfo{year}{2015}\natexlab{}.
\newblock \showarticletitle{Inner Speech: Development, Cognitive Functions,
  Phenomenology, and Neurobiology}.
\newblock In \bibinfo{booktitle}{{\em Psychological Bulletin, American
  Psychological Association}}.
\newblock


\bibitem[\protect\citeauthoryear{Andreas}{Andreas}{2019}]%
        {DBLP:journals/corr/abs-1902-07181}
\bibfield{author}{\bibinfo{person}{Jacob Andreas}.}
  \bibinfo{year}{2019}\natexlab{}.
\newblock \showarticletitle{Measuring Compositionality in Representation
  Learning}. In \bibinfo{booktitle}{{\em International Conference on Learning
  Representations}}.
\newblock
\showURL{%
\url{https://openreview.net/forum?id=HJz05o0qK7}}


\bibitem[\protect\citeauthoryear{Baars}{Baars}{2017}]%
        {doi:10.1002/9781119132363.ch16}
\bibfield{author}{\bibinfo{person}{Bernard~J. Baars}.}
  \bibinfo{year}{2017}\natexlab{}.
\newblock \bibinfo{booktitle}{{\em The Global Workspace Theory of
  Consciousness}}.
\newblock \bibinfo{publisher}{John Wiley \& Sons, Ltd}, Chapter~16,
  \bibinfo{pages}{227--242}.
\newblock
\showISBNx{9781119132363}
\showDOI{%
\url{https://doi.org/10.1002/9781119132363.ch16}}
\showeprint{https://onlinelibrary.wiley.com/doi/pdf/10.1002/9781119132363.ch16}


\bibitem[\protect\citeauthoryear{Bacharach}{Bacharach}{1991}]%
        {Bacharach1991}
\bibfield{author}{\bibinfo{person}{Michael Bacharach}.}
  \bibinfo{year}{1991}\natexlab{}.
\newblock \bibinfo{booktitle}{{\em Zero-sum Games}}.
\newblock \bibinfo{publisher}{Palgrave Macmillan UK},
  \bibinfo{address}{London}, \bibinfo{pages}{727--731}.
\newblock
\showISBNx{978-1-349-21315-3}
\showDOI{%
\url{https://doi.org/10.1007/978-1-349-21315-3_100}}


\bibitem[\protect\citeauthoryear{Baddeley and Hitch}{Baddeley and
  Hitch}{1974}]%
        {BADDELEY197447}
\bibfield{author}{\bibinfo{person}{Alan~D. Baddeley} {and}
  \bibinfo{person}{Graham Hitch}.} \bibinfo{year}{1974}\natexlab{}.
\newblock \showarticletitle{Working Memory}.
\newblock \bibinfo{series}{Psychology of Learning and Motivation},
  Vol.~\bibinfo{volume}{8}. \bibinfo{publisher}{Academic Press},
  \bibinfo{pages}{47 -- 89}.
\newblock
\showISSN{0079-7421}
\showDOI{%
\url{https://doi.org/10.1016/S0079-7421(08)60452-1}}


\bibitem[\protect\citeauthoryear{Baligh and Richartz}{Baligh and
  Richartz}{1967}]%
        {doi:10.1177/002224376700400209}
\bibfield{author}{\bibinfo{person}{Helmy~H. Baligh} {and}
  \bibinfo{person}{Leon~E. Richartz}.} \bibinfo{year}{1967}\natexlab{}.
\newblock \showarticletitle{Variable-Sum Game Models of Marketing Problems}.
\newblock \bibinfo{journal}{{\em Journal of Marketing Research\/}}
  \bibinfo{volume}{4}, \bibinfo{number}{2} (\bibinfo{year}{1967}),
  \bibinfo{pages}{173--183}.
\newblock
\showDOI{%
\url{https://doi.org/10.1177/002224376700400209}}
\showeprint{https://doi.org/10.1177/002224376700400209}


\bibitem[\protect\citeauthoryear{Bassok}{Bassok}{2012}]%
        {doi:10.1177/0895904811428973}
\bibfield{author}{\bibinfo{person}{Daphna Bassok}.}
  \bibinfo{year}{2012}\natexlab{}.
\newblock \showarticletitle{Competition or Collaboration?: Head Start
  Enrollment During the Rapid Expansion of State Pre-kindergarten}.
\newblock \bibinfo{journal}{{\em Educational Policy\/}} \bibinfo{volume}{26},
  \bibinfo{number}{1} (\bibinfo{year}{2012}), \bibinfo{pages}{96--116}.
\newblock
\showDOI{%
\url{https://doi.org/10.1177/0895904811428973}}
\showeprint{https://doi.org/10.1177/0895904811428973}


\bibitem[\protect\citeauthoryear{Bogin, Geva, and Berant}{Bogin
  et~al\mbox{.}}{2018}]%
        {DBLP:journals/corr/abs-1809-00549}
\bibfield{author}{\bibinfo{person}{Ben Bogin}, \bibinfo{person}{Mor Geva},
  {and} \bibinfo{person}{Jonathan Berant}.} \bibinfo{year}{2018}\natexlab{}.
\newblock \showarticletitle{Emergence of Communication in an Interactive World
  with Consistent Speakers}.
\newblock \bibinfo{journal}{{\em CoRR\/}}  \bibinfo{volume}{abs/1809.00549}
  (\bibinfo{year}{2018}).
\newblock
\showeprint[arxiv]{1809.00549}
\showURL{%
\url{http://arxiv.org/abs/1809.00549}}


\bibitem[\protect\citeauthoryear{Bordes and Weston}{Bordes and Weston}{2016}]%
        {DBLP:journals/corr/BordesW16}
\bibfield{author}{\bibinfo{person}{Antoine Bordes} {and} \bibinfo{person}{Jason
  Weston}.} \bibinfo{year}{2016}\natexlab{}.
\newblock \showarticletitle{Learning End-to-End Goal-Oriented Dialog}.
\newblock \bibinfo{journal}{{\em CoRR\/}}  \bibinfo{volume}{abs/1605.07683}
  (\bibinfo{year}{2016}).
\newblock
\showeprint[arxiv]{1605.07683}
\showURL{%
\url{http://arxiv.org/abs/1605.07683}}


\bibitem[\protect\citeauthoryear{Bouchacourt and Baroni}{Bouchacourt and
  Baroni}{2018}]%
        {DBLP:journals/corr/abs-1808-10696}
\bibfield{author}{\bibinfo{person}{Diane Bouchacourt} {and}
  \bibinfo{person}{Marco Baroni}.} \bibinfo{year}{2018}\natexlab{}.
\newblock \showarticletitle{How agents see things: On visual representations in
  an emergent language game}.
\newblock \bibinfo{journal}{{\em CoRR\/}}  \bibinfo{volume}{abs/1808.10696}
  (\bibinfo{year}{2018}).
\newblock
\showeprint[arxiv]{1808.10696}
\showURL{%
\url{http://arxiv.org/abs/1808.10696}}


\bibitem[\protect\citeauthoryear{Chomsky}{Chomsky}{1957}]%
        {reason:Chomsky57a}
\bibfield{author}{\bibinfo{person}{Noam Chomsky}.}
  \bibinfo{year}{1957}\natexlab{}.
\newblock \bibinfo{booktitle}{{\em Syntactic Structures}}.
\newblock \bibinfo{publisher}{Mouton and Co.}, \bibinfo{address}{The Hague}.
\newblock


\bibitem[\protect\citeauthoryear{Christiansen and Loeschcke}{Christiansen and
  Loeschcke}{1990}]%
        {Christiansen1990}
\bibfield{author}{\bibinfo{person}{F.~B. Christiansen} {and}
  \bibinfo{person}{V. Loeschcke}.} \bibinfo{year}{1990}\natexlab{}.
\newblock \bibinfo{booktitle}{{\em Evolution and Competition}}.
\newblock \bibinfo{publisher}{Springer Berlin Heidelberg},
  \bibinfo{address}{Berlin, Heidelberg}, \bibinfo{pages}{367--394}.
\newblock
\showISBNx{978-3-642-74474-7}
\showDOI{%
\url{https://doi.org/10.1007/978-3-642-74474-7_13}}


\bibitem[\protect\citeauthoryear{Cogswell, Lu, Lee, Parikh, and Batra}{Cogswell
  et~al\mbox{.}}{2019}]%
        {DBLP:journals/corr/abs-1904-09067}
\bibfield{author}{\bibinfo{person}{Michael Cogswell}, \bibinfo{person}{Jiasen
  Lu}, \bibinfo{person}{Stefan Lee}, \bibinfo{person}{Devi Parikh}, {and}
  \bibinfo{person}{Dhruv Batra}.} \bibinfo{year}{2019}\natexlab{}.
\newblock \showarticletitle{Emergence of Compositional Language with Deep
  Generational Transmission}.
\newblock \bibinfo{journal}{{\em CoRR\/}}  \bibinfo{volume}{abs/1904.09067}
  (\bibinfo{year}{2019}).
\newblock
\showeprint[arxiv]{1904.09067}
\showURL{%
\url{http://arxiv.org/abs/1904.09067}}


\bibitem[\protect\citeauthoryear{Das, Datta, Gkioxari, Lee, Parikh, and
  Batra}{Das et~al\mbox{.}}{2018}]%
        {DBLP:journals/corr/abs-1711-11543}
\bibfield{author}{\bibinfo{person}{Abhishek Das}, \bibinfo{person}{Samyak
  Datta}, \bibinfo{person}{Georgia Gkioxari}, \bibinfo{person}{Stefan Lee},
  \bibinfo{person}{Devi Parikh}, {and} \bibinfo{person}{Dhruv Batra}.}
  \bibinfo{year}{2018}\natexlab{}.
\newblock \showarticletitle{{E}mbodied {Q}uestion {A}nswering}. In
  \bibinfo{booktitle}{{\em Proceedings of the IEEE Conference on Computer
  Vision and Pattern Recognition (CVPR)}}.
\newblock


\bibitem[\protect\citeauthoryear{DiMenichi and Tricomi}{DiMenichi and
  Tricomi}{2015}]%
        {compete}
\bibfield{author}{\bibinfo{person}{Brynne~C. DiMenichi} {and}
  \bibinfo{person}{Elizabeth Tricomi}.} \bibinfo{year}{2015}\natexlab{}.
\newblock \showarticletitle{The power of competition: Effects of social
  motivation on attention, sustained physical effort, and learning}.
\newblock \bibinfo{journal}{{\em Frontiers in Psychology\/}}
  (\bibinfo{year}{2015}).
\newblock


\bibitem[\protect\citeauthoryear{Glorot, Bordes, and Bengio}{Glorot
  et~al\mbox{.}}{2011}]%
        {Glorot:2011:DAL:3104482.3104547}
\bibfield{author}{\bibinfo{person}{Xavier Glorot}, \bibinfo{person}{Antoine
  Bordes}, {and} \bibinfo{person}{Yoshua Bengio}.}
  \bibinfo{year}{2011}\natexlab{}.
\newblock \showarticletitle{Domain Adaptation for Large-scale Sentiment
  Classification: A Deep Learning Approach}. In \bibinfo{booktitle}{{\em
  ICML}}.
\newblock


\bibitem[\protect\citeauthoryear{{Grossberg}}{{Grossberg}}{2013}]%
        {6706709}
\bibfield{author}{\bibinfo{person}{S. {Grossberg}}.}
  \bibinfo{year}{2013}\natexlab{}.
\newblock \showarticletitle{Behavioral economics and neuroeconomics:
  Cooperation, competition, preference, and decision making}. In
  \bibinfo{booktitle}{{\em The 2013 International Joint Conference on Neural
  Networks (IJCNN)}}. \bibinfo{pages}{1--5}.
\newblock
\showISSN{2161-4393}
\showDOI{%
\url{https://doi.org/10.1109/IJCNN.2013.6706709}}


\bibitem[\protect\citeauthoryear{Grze\'{s}}{Grze\'{s}}{2017}]%
        {Grzes:2017:RSE:3091125.3091208}
\bibfield{author}{\bibinfo{person}{Marek Grze\'{s}}.}
  \bibinfo{year}{2017}\natexlab{}.
\newblock \showarticletitle{Reward Shaping in Episodic Reinforcement Learning}.
  In \bibinfo{booktitle}{{\em AAMAS}}.
\newblock


\bibitem[\protect\citeauthoryear{Grze\'{s} and Kudenko}{Grze\'{s} and
  Kudenko}{2008}]%
        {Grzes:2008:MRL:1429293.1429335}
\bibfield{author}{\bibinfo{person}{Marek Grze\'{s}} {and}
  \bibinfo{person}{Daniel Kudenko}.} \bibinfo{year}{2008}\natexlab{}.
\newblock \showarticletitle{Multigrid Reinforcement Learning with Reward
  Shaping}. In \bibinfo{booktitle}{{\em ICANN}}.
\newblock


\bibitem[\protect\citeauthoryear{Hochreiter and Schmidhuber}{Hochreiter and
  Schmidhuber}{1997}]%
        {hochreiter1997long}
\bibfield{author}{\bibinfo{person}{Sepp Hochreiter} {and}
  \bibinfo{person}{J{\"u}rgen Schmidhuber}.} \bibinfo{year}{1997}\natexlab{}.
\newblock \showarticletitle{Long short-term memory}.
\newblock \bibinfo{journal}{{\em Neural computation\/}} \bibinfo{volume}{9},
  \bibinfo{number}{8} (\bibinfo{year}{1997}), \bibinfo{pages}{1735--1780}.
\newblock


\bibitem[\protect\citeauthoryear{Jain and Medsker}{Jain and Medsker}{1999}]%
        {Jain:1999:RNN:553011}
\bibfield{author}{\bibinfo{person}{L.~C. Jain} {and} \bibinfo{person}{L.~R.
  Medsker}.} \bibinfo{year}{1999}\natexlab{}.
\newblock \bibinfo{booktitle}{{\em Recurrent Neural Networks: Design and
  Applications\/} (\bibinfo{edition}{1st} ed.)}.
\newblock \bibinfo{publisher}{CRC Press, Inc.}, \bibinfo{address}{Boca Raton,
  FL, USA}.
\newblock
\showISBNx{0849371813}


\bibitem[\protect\citeauthoryear{Jaques, Lazaridou, Hughes,
  G{\"{u}}l{\c{c}}ehre, Ortega, Strouse, Leibo, and de~Freitas}{Jaques
  et~al\mbox{.}}{2019}]%
        {jaques2019intrinsic}
\bibfield{author}{\bibinfo{person}{Natasha Jaques}, \bibinfo{person}{Angeliki
  Lazaridou}, \bibinfo{person}{Edward Hughes}, \bibinfo{person}{{\c{C}}aglar
  G{\"{u}}l{\c{c}}ehre}, \bibinfo{person}{Pedro~A. Ortega}, \bibinfo{person}{DJ
  Strouse}, \bibinfo{person}{Joel~Z. Leibo}, {and} \bibinfo{person}{Nando de
  Freitas}.} \bibinfo{year}{2019}\natexlab{}.
\newblock \showarticletitle{Social Influence as Intrinsic Motivation for
  Multi-Agent Deep Reinforcement Learning}. In \bibinfo{booktitle}{{\em
  Proceedings of the 36th International Conference on Machine Learning, {ICML}
  2019, 9-15 June 2019, Long Beach, California, {USA}}} {\em
  (\bibinfo{series}{Proceedings of Machine Learning Research})},
  \bibfield{editor}{\bibinfo{person}{Kamalika Chaudhuri} {and}
  \bibinfo{person}{Ruslan Salakhutdinov}} (Eds.), Vol.~\bibinfo{volume}{97}.
  \bibinfo{publisher}{{PMLR}}, \bibinfo{pages}{3040--3049}.
\newblock
\showURL{%
\url{http://proceedings.mlr.press/v97/jaques19a.html}}


\bibitem[\protect\citeauthoryear{Kirby, Cornish, and Smith}{Kirby
  et~al\mbox{.}}{2008}]%
        {Kirby10681}
\bibfield{author}{\bibinfo{person}{Simon Kirby}, \bibinfo{person}{Hannah
  Cornish}, {and} \bibinfo{person}{Kenny Smith}.}
  \bibinfo{year}{2008}\natexlab{}.
\newblock \showarticletitle{Cumulative cultural evolution in the laboratory: An
  experimental approach to the origins of structure in human language}.
\newblock \bibinfo{journal}{{\em Proceedings of the National Academy of
  Sciences\/}} \bibinfo{volume}{105}, \bibinfo{number}{31}
  (\bibinfo{year}{2008}), \bibinfo{pages}{10681--10686}.
\newblock
\showISSN{0027-8424}
\showDOI{%
\url{https://doi.org/10.1073/pnas.0707835105}}
\showeprint{https://www.pnas.org/content/105/31/10681.full.pdf}


\bibitem[\protect\citeauthoryear{Kirby, Tamariz, Cornish, and Smith}{Kirby
  et~al\mbox{.}}{2015}]%
        {Kirby2015CompressionAC}
\bibfield{author}{\bibinfo{person}{Simon Kirby}, \bibinfo{person}{Monica
  Tamariz}, \bibinfo{person}{Hannah Cornish}, {and} \bibinfo{person}{Kenny
  Smith}.} \bibinfo{year}{2015}\natexlab{}.
\newblock \showarticletitle{Compression and communication in the cultural
  evolution of linguistic structure}.
\newblock \bibinfo{journal}{{\em Cognition\/}}  \bibinfo{volume}{141}
  (\bibinfo{year}{2015}), \bibinfo{pages}{87--102}.
\newblock


\bibitem[\protect\citeauthoryear{Kottur, Moura, Lee, and Batra}{Kottur
  et~al\mbox{.}}{2017}]%
        {DBLP:journals/corr/KotturMLB17}
\bibfield{author}{\bibinfo{person}{Satwik Kottur}, \bibinfo{person}{Jos{\'e}
  Moura}, \bibinfo{person}{Stefan Lee}, {and} \bibinfo{person}{Dhruv Batra}.}
  \bibinfo{year}{2017}\natexlab{}.
\newblock \showarticletitle{Natural Language Does Not Emerge {`}Naturally{'} in
  Multi-Agent Dialog}. In \bibinfo{booktitle}{{\em Proceedings of the 2017
  Conference on Empirical Methods in Natural Language Processing}}.
  \bibinfo{publisher}{Association for Computational Linguistics},
  \bibinfo{address}{Copenhagen, Denmark}, \bibinfo{pages}{2962--2967}.
\newblock
\showDOI{%
\url{https://doi.org/10.18653/v1/D17-1321}}


\bibitem[\protect\citeauthoryear{Lazaridou, Hermann, Tuyls, and
  Clark}{Lazaridou et~al\mbox{.}}{2018}]%
        {lazaridou2018emergence}
\bibfield{author}{\bibinfo{person}{Angeliki Lazaridou},
  \bibinfo{person}{Karl~Moritz Hermann}, \bibinfo{person}{Karl Tuyls}, {and}
  \bibinfo{person}{Stephen Clark}.} \bibinfo{year}{2018}\natexlab{}.
\newblock \showarticletitle{Emergence of Linguistic Communication from
  Referential Games with Symbolic and Pixel Input}. In \bibinfo{booktitle}{{\em
  International Conference on Learning Representations}}.
\newblock
\showURL{%
\url{https://openreview.net/forum?id=HJGv1Z-AW}}


\bibitem[\protect\citeauthoryear{Lazaridou, Peysakhovich, and Baroni}{Lazaridou
  et~al\mbox{.}}{2017}]%
        {DBLP:journals/corr/LazaridouPB16b}
\bibfield{author}{\bibinfo{person}{Angeliki Lazaridou},
  \bibinfo{person}{Alexander Peysakhovich}, {and} \bibinfo{person}{Marco
  Baroni}.} \bibinfo{year}{2017}\natexlab{}.
\newblock \showarticletitle{Multi-Agent Cooperation and the Emergence of
  (Natural) Language}. In \bibinfo{booktitle}{{\em 5th International Conference
  on Learning Representations, {ICLR} 2017, Toulon, France, April 24-26, 2017,
  Conference Track Proceedings}}. \bibinfo{publisher}{OpenReview.net}.
\newblock
\showURL{%
\url{https://openreview.net/forum?id=Hk8N3Sclg}}


\bibitem[\protect\citeauthoryear{Leigh~Jr}{Leigh~Jr}{2010}]%
        {doi:10.1111/j.1420-9101.2010.02114.x}
\bibfield{author}{\bibinfo{person}{E.~G. Leigh~Jr}.}
  \bibinfo{year}{2010}\natexlab{}.
\newblock \showarticletitle{The evolution of mutualism}.
\newblock \bibinfo{journal}{{\em Journal of Evolutionary Biology\/}}
  \bibinfo{volume}{23}, \bibinfo{number}{12} (\bibinfo{year}{2010}),
  \bibinfo{pages}{2507--2528}.
\newblock
\showDOI{%
\url{https://doi.org/10.1111/j.1420-9101.2010.02114.x}}
\showeprint{https://onlinelibrary.wiley.com/doi/pdf/10.1111/j.1420-9101.2010.02114.x}


\bibitem[\protect\citeauthoryear{Liang, Lim, Tsai, Salakhutdinov, and
  Morency}{Liang et~al\mbox{.}}{2019}]%
        {DBLP:journals/corr/abs-1906-02125}
\bibfield{author}{\bibinfo{person}{Paul~Pu Liang}, \bibinfo{person}{Yao~Chong
  Lim}, \bibinfo{person}{Yao-Hung~Hubert Tsai}, \bibinfo{person}{Ruslan
  Salakhutdinov}, {and} \bibinfo{person}{Louis-Philippe Morency}.}
  \bibinfo{year}{2019}\natexlab{}.
\newblock \showarticletitle{Strong and Simple Baselines for Multimodal
  Utterance Embeddings}. In \bibinfo{booktitle}{{\em Proceedings of the 2019
  Conference of the North {A}merican Chapter of the Association for
  Computational Linguistics: Human Language Technologies, Volume 1 (Long and
  Short Papers)}}. \bibinfo{publisher}{Association for Computational
  Linguistics}, \bibinfo{address}{Minneapolis, Minnesota},
  \bibinfo{pages}{2599--2609}.
\newblock
\showDOI{%
\url{https://doi.org/10.18653/v1/N19-1267}}


\bibitem[\protect\citeauthoryear{Liang, Liu, Bagher~Zadeh, and Morency}{Liang
  et~al\mbox{.}}{2018}]%
        {multistage}
\bibfield{author}{\bibinfo{person}{Paul~Pu Liang}, \bibinfo{person}{Ziyin Liu},
  \bibinfo{person}{AmirAli Bagher~Zadeh}, {and} \bibinfo{person}{Louis-Philippe
  Morency}.} \bibinfo{year}{2018}\natexlab{}.
\newblock \showarticletitle{Multimodal Language Analysis with Recurrent
  Multistage Fusion}. In \bibinfo{booktitle}{{\em Proceedings of the 2018
  Conference on Empirical Methods in Natural Language Processing}}.
  \bibinfo{publisher}{Association for Computational Linguistics},
  \bibinfo{address}{Brussels, Belgium}, \bibinfo{pages}{150--161}.
\newblock
\showDOI{%
\url{https://doi.org/10.18653/v1/D18-1014}}


\bibitem[\protect\citeauthoryear{Lowe, Foerster, Boureau, Pineau, and
  Dauphin}{Lowe et~al\mbox{.}}{2019}]%
        {DBLP:journals/corr/abs-1903-05168}
\bibfield{author}{\bibinfo{person}{Ryan Lowe}, \bibinfo{person}{Jakob
  Foerster}, \bibinfo{person}{Y-Lan Boureau}, \bibinfo{person}{Joelle Pineau},
  {and} \bibinfo{person}{Yann Dauphin}.} \bibinfo{year}{2019}\natexlab{}.
\newblock \showarticletitle{On the Pitfalls of Measuring Emergent
  Communication}. In \bibinfo{booktitle}{{\em Proceedings of the 18th
  International Conference on Autonomous Agents and MultiAgent Systems}} {\em
  (\bibinfo{series}{AAMAS ’19})}. \bibinfo{publisher}{International
  Foundation for Autonomous Agents and Multiagent Systems},
  \bibinfo{address}{Richland, SC}, \bibinfo{pages}{693–701}.
\newblock
\showISBNx{9781450363099}


\bibitem[\protect\citeauthoryear{Lowe, Wu, Tamar, Harb, Abbeel, and
  Mordatch}{Lowe et~al\mbox{.}}{2017}]%
        {DBLP:journals/corr/LoweWTHAM17}
\bibfield{author}{\bibinfo{person}{Ryan Lowe}, \bibinfo{person}{Yi Wu},
  \bibinfo{person}{Aviv Tamar}, \bibinfo{person}{Jean Harb},
  \bibinfo{person}{Pieter Abbeel}, {and} \bibinfo{person}{Igor Mordatch}.}
  \bibinfo{year}{2017}\natexlab{}.
\newblock \showarticletitle{Multi-Agent Actor-Critic for Mixed
  Cooperative-Competitive Environments}. In \bibinfo{booktitle}{{\em
  Proceedings of the 31st International Conference on Neural Information
  Processing Systems}} {\em (\bibinfo{series}{NIPS’17})}.
  \bibinfo{publisher}{Curran Associates Inc.}, \bibinfo{address}{Red Hook, NY,
  USA}, \bibinfo{pages}{6382–6393}.
\newblock
\showISBNx{9781510860964}


\bibitem[\protect\citeauthoryear{Mobbs, Hassabis, Yu, Chu, Rushworth, Boorman,
  and Dalgleish}{Mobbs et~al\mbox{.}}{2013}]%
        {Mobbs9866}
\bibfield{author}{\bibinfo{person}{Dean Mobbs}, \bibinfo{person}{Demis
  Hassabis}, \bibinfo{person}{Rongjun Yu}, \bibinfo{person}{Carlton Chu},
  \bibinfo{person}{Matthew Rushworth}, \bibinfo{person}{Erie Boorman}, {and}
  \bibinfo{person}{Tim Dalgleish}.} \bibinfo{year}{2013}\natexlab{}.
\newblock \showarticletitle{Foraging under Competition: The Neural Basis of
  Input-Matching in Humans}.
\newblock \bibinfo{journal}{{\em Journal of Neuroscience\/}}
  \bibinfo{volume}{33}, \bibinfo{number}{23} (\bibinfo{year}{2013}),
  \bibinfo{pages}{9866--9872}.
\newblock
\showISSN{0270-6474}
\showDOI{%
\url{https://doi.org/10.1523/JNEUROSCI.2238-12.2013}}
\showeprint{http://www.jneurosci.org/content/33/23/9866.full.pdf}


\bibitem[\protect\citeauthoryear{Nowak and Krakauer}{Nowak and
  Krakauer}{1999}]%
        {Nowak8028}
\bibfield{author}{\bibinfo{person}{Martin~A. Nowak} {and}
  \bibinfo{person}{David~C. Krakauer}.} \bibinfo{year}{1999}\natexlab{}.
\newblock \showarticletitle{The evolution of language}.
\newblock \bibinfo{journal}{{\em Proceedings of the National Academy of
  Sciences\/}} \bibinfo{volume}{96}, \bibinfo{number}{14}
  (\bibinfo{year}{1999}), \bibinfo{pages}{8028--8033}.
\newblock
\showISSN{0027-8424}
\showDOI{%
\url{https://doi.org/10.1073/pnas.96.14.8028}}
\showeprint{https://www.pnas.org/content/96/14/8028.full.pdf}


\bibitem[\protect\citeauthoryear{Nowak, Plotkin, and Jansen}{Nowak
  et~al\mbox{.}}{2000}]%
        {Nowak2000TheEO}
\bibfield{author}{\bibinfo{person}{Martin~A. Nowak}, \bibinfo{person}{Joshua~B.
  Plotkin}, {and} \bibinfo{person}{Vincent A~A Jansen}.}
  \bibinfo{year}{2000}\natexlab{}.
\newblock \showarticletitle{The evolution of syntactic communication}.
\newblock \bibinfo{journal}{{\em Nature\/}}  \bibinfo{volume}{404}
  (\bibinfo{year}{2000}), \bibinfo{pages}{495--498}.
\newblock


\bibitem[\protect\citeauthoryear{Pekkonen, Ketola, and Laakso}{Pekkonen
  et~al\mbox{.}}{2013}]%
        {fasfas}
\bibfield{author}{\bibinfo{person}{Minna Pekkonen}, \bibinfo{person}{Tarmo
  Ketola}, {and} \bibinfo{person}{Jouni~T Laakso}.}
  \bibinfo{year}{2013}\natexlab{}.
\newblock \showarticletitle{Resource availability and competition shape the
  evolution of survival and growth ability in a bacterial community.}
\newblock \bibinfo{journal}{{\em PLoS One\/}} \bibinfo{volume}{8},
  \bibinfo{number}{9} (\bibinfo{year}{2013}).
\newblock


\bibitem[\protect\citeauthoryear{Pepitone}{Pepitone}{1985}]%
        {Pepitone1985}
\bibfield{author}{\bibinfo{person}{Emmy~A. Pepitone}.}
  \bibinfo{year}{1985}\natexlab{}.
\newblock \bibinfo{booktitle}{{\em Children in Cooperation and Competition}}.
\newblock \bibinfo{publisher}{Springer US}, \bibinfo{address}{Boston, MA},
  \bibinfo{pages}{17--65}.
\newblock
\showISBNx{978-1-4899-3650-9}
\showDOI{%
\url{https://doi.org/10.1007/978-1-4899-3650-9_2}}


\bibitem[\protect\citeauthoryear{Pham, Liang, Manzini, Morency, and
  P{\'{o}}czos}{Pham et~al\mbox{.}}{2019}]%
        {DBLP:conf/aaai/PhamLMMP19}
\bibfield{author}{\bibinfo{person}{Hai Pham}, \bibinfo{person}{Paul~Pu Liang},
  \bibinfo{person}{Thomas Manzini}, \bibinfo{person}{Louis{-}Philippe Morency},
  {and} \bibinfo{person}{Barnab{\'{a}}s P{\'{o}}czos}.}
  \bibinfo{year}{2019}\natexlab{}.
\newblock \showarticletitle{Found in Translation: Learning Robust Joint
  Representations by Cyclic Translations between Modalities}. In
  \bibinfo{booktitle}{{\em The Thirty-Third {AAAI} Conference on Artificial
  Intelligence, {AAAI} 2019}}. \bibinfo{pages}{6892--6899}.
\newblock
\showDOI{%
\url{https://doi.org/10.1609/aaai.v33i01.33016892}}


\bibitem[\protect\citeauthoryear{Quionero-Candela, Sugiyama, Schwaighofer, and
  Lawrence}{Quionero-Candela et~al\mbox{.}}{2009}]%
        {Quionero-Candela:2009:DSM:1462129}
\bibfield{author}{\bibinfo{person}{Joaquin Quionero-Candela},
  \bibinfo{person}{Masashi Sugiyama}, \bibinfo{person}{Anton Schwaighofer},
  {and} \bibinfo{person}{Neil~D. Lawrence}.} \bibinfo{year}{2009}\natexlab{}.
\newblock \bibinfo{booktitle}{{\em Dataset Shift in Machine Learning}}.
\newblock \bibinfo{publisher}{The MIT Press}.
\newblock
\showISBNx{0262170051, 9780262170055}


\bibitem[\protect\citeauthoryear{Read}{Read}{2012}]%
        {doi:10.1080/2158379X.2012.659865}
\bibfield{author}{\bibinfo{person}{James~H. Read}.}
  \bibinfo{year}{2012}\natexlab{}.
\newblock \showarticletitle{Is power zero-sum or variable-sum? Old arguments
  and new beginnings}.
\newblock \bibinfo{journal}{{\em Journal of Political Power\/}}
  (\bibinfo{year}{2012}).
\newblock
\showDOI{%
\url{https://doi.org/10.1080/2158379X.2012.659865}}
\showeprint{https://doi.org/10.1080/2158379X.2012.659865}


\bibitem[\protect\citeauthoryear{Smith, Kirby, and Brighton}{Smith
  et~al\mbox{.}}{2003}]%
        {Smith:2003:ILF:963725.963729}
\bibfield{author}{\bibinfo{person}{Kenny Smith}, \bibinfo{person}{Simon Kirby},
  {and} \bibinfo{person}{Henry Brighton}.} \bibinfo{year}{2003}\natexlab{}.
\newblock \showarticletitle{Iterated Learning: A Framework for the Emergence of
  Language}.
\newblock \bibinfo{journal}{{\em Artif. Life\/}} \bibinfo{volume}{9},
  \bibinfo{number}{4} (\bibinfo{date}{Sept.} \bibinfo{year}{2003}),
  \bibinfo{pages}{371--386}.
\newblock
\showISSN{1064-5462}
\showDOI{%
\url{https://doi.org/10.1162/106454603322694825}}


\bibitem[\protect\citeauthoryear{Tamariz and Kirby}{Tamariz and Kirby}{2015}]%
        {articleaaa122}
\bibfield{author}{\bibinfo{person}{Monica Tamariz} {and} \bibinfo{person}{Simon
  Kirby}.} \bibinfo{year}{2015}\natexlab{}.
\newblock \showarticletitle{The Cultural Evolution of Language}.
\newblock \bibinfo{journal}{{\em Current Opinion in Psychology\/}}
  \bibinfo{volume}{8} (\bibinfo{date}{09} \bibinfo{year}{2015}).
\newblock
\showDOI{%
\url{https://doi.org/10.1016/j.copsyc.2015.09.003}}


\bibitem[\protect\citeauthoryear{Vogt}{Vogt}{2005}]%
        {Vogt2005TheEO}
\bibfield{author}{\bibinfo{person}{Paul Vogt}.}
  \bibinfo{year}{2005}\natexlab{}.
\newblock \showarticletitle{The emergence of compositional structures in
  perceptually grounded language games}.
\newblock \bibinfo{journal}{{\em Artif. Intell.\/}}  \bibinfo{volume}{167}
  (\bibinfo{year}{2005}), \bibinfo{pages}{206--242}.
\newblock


\bibitem[\protect\citeauthoryear{Williams}{Williams}{1992}]%
        {williams1992simple}
\bibfield{author}{\bibinfo{person}{Ronald~J. Williams}.}
  \bibinfo{year}{1992}\natexlab{}.
\newblock \showarticletitle{Simple Statistical Gradient-Following Algorithms
  for Connectionist Reinforcement Learning}.
\newblock \bibinfo{journal}{{\em Mach. Learn.\/}} \bibinfo{volume}{8},
  \bibinfo{number}{3-4} (\bibinfo{date}{May} \bibinfo{year}{1992}),
  \bibinfo{pages}{229--256}.
\newblock
\showISSN{0885-6125}
\showDOI{%
\url{https://doi.org/10.1007/BF00992696}}


\end{thebibliography}

\clearpage
\appendix

\section*{Appendix}

\section{Modeling the Agents}
\label{model}

The \ttnc \ game begins with two target instances $I_{(1)}$ and $I_{(2)}$ presented to \Qbot$_{(1)}$ and \Qbot$_{(2)}$ respectively, and two tasks $G_{(1)}$ and $G_{(2)}$ presented to \Abot$_{(1)}$ and \Abot$_{(2)}$ respectively. Within a team, we largely follow the setting by~\citet{DBLP:journals/corr/KotturMLB17}. A team consists of agents \Qbot \ and \Abot \ cooperating in a partially observable world to solve task $G$ given instance $I$. We use lower case characters (e.g. $s_t^Q$) to denote the token symbol and upper case $S_t^Q$ to denote the corresponding representation. We use subscripts to index the rounds and subscripts in parenthesis to index which team the pair of agents belong to (i.e. ${s_t^Q}_{(1)}$). We drop the team subscript if it is clear from the context (i.e. same team).

\textbf{Base States and Actions:} Each agent observes its specific input (task $G$ for \Qbot \ and instance instance $I$ for \Abot) and the output of the other agent in the same team. At the beginning of round $t$, \Qbot \ observes state $s^Q_t = [G, q_1, a_1, \dots, q_{t-1}, a_{t-1}]$ and utters some token $q_t \in V_Q$. \Abot \ observes the history and this new utterance as state $s^A_t = [F, q_1, a_1, \dots, q_{t-1},a_{t-1},q_t]$ and utters $a_t \in V_A$. Note, the two teams share vocabularies (i.e. $V_Q$ is shared by \Qbot$_{(1)}$ and \Qbot$_{(2)}$). At the final round, \Qbot \ predicts a pair of attribute values $\hat{w}^G = (\hat{w}^G_1, \hat{w}^G_2)$ to solve the task. \Qbot \ and \Abot \ are modeled as \textbf{stochastic policies} $\pi_Q(q_t{\mid}s^Q_t; \theta_Q)$ and $\pi_A(a_t{\mid}s^A_t; \theta_A)$ implemented as recurrent networks~\cite{Jain:1999:RNN:553011,hochreiter1997long}. \Qbot \ is modeled with three modules -- speaking, listening, and prediction. Given task $G$, \Qbot \ stores an initial state $S^Q_{t-1}$ from which it conditionally generates output utterances $q_t\in V_Q$. $S^Q_{t-1}$ is updated using answers $a_t$ from \Abot \ and is used to make a prediction $\hat{w}_G$ in the final round. \Abot \ is modeled with two modules -- speaking and listening. \Abot \ encodes instance $I$ into its initial state $S_t^A$ from which it conditionally generates output utterances $a_t\in V_A$. $S_t^A$ is updated using questions $a_t$ from \Qbot. \Qbot \ and \Abot \ receive an identical \textbf{base reward} of $R$ if \Qbot's prediction $\hat{w}^G$ matches ground truth $w^G$ and a negative reward of $-10R$ otherwise. $R$ is a hyperparamter which affects the rate of convergence. To train these agents, we update policy parameters $\theta_Q$ and $\theta_A$ using the popular REINFORCE policy gradient algorithm~\cite{williams1992simple}.

\begin{table}[t]
\fontsize{8.5}{11}\selectfont
\centering
\setlength\tabcolsep{3.5pt}
\begin{tabular}{c | c}
\Xhline{3\arrayrulewidth}
Parameter & Value \\
\Xhline{0.5\arrayrulewidth}
attribute embeddings size & 20 \\
instance embedding size & 60 \\
$R$ & 100 \\
$|V_Q|$ & 3 \\
$|V_A|$ & 4 \\
$\rho$ & 0.5 \\
batch size & 1000 \\
LSTM dimension & 50 \\
episodes & 1000 \\
max epochs & 50000 \\
learning rate & 0.01 \\
gradient clipping & [-5.0,+5.0] \\
optimizer & Adam \\
num repeats & 10 \\
\Xhline{3\arrayrulewidth}
\end{tabular}
\caption{Table of hyperparameters.}
\label{config}
\end{table}

\section{Implementation Details}

Table~\ref{config} is the list of hyperparameters used. We train all models using the same set of hyperparameters and only modify the rewards and information being shared among agents. All reported results were averaged over 10 runs with randomly initialized random seeds.

In order to report results on the Instantaneous Coordination (IC) metric, we compute the mutual information across the following: within team 1, the mutual information between \Abot's messages and \Qbot's first guess, and \Abot's messages and \Qbot's second guess. We average these number to obtain an aggregated IC score for team 1 before repeating the procedure for team 2.

\end{document}